%% file: neurips_2022.tex
\documentclass{article}

\usepackage[final, nonatbib]{neurips_2022}

\usepackage[utf8]{inputenc} 
\usepackage[T1]{fontenc}    
\usepackage[hidelinks]{hyperref}       
\usepackage{url}            
\usepackage{booktabs}       
\usepackage{amsfonts}       
\usepackage{nicefrac}       
\usepackage{microtype}      
\usepackage[table]{xcolor}         

\usepackage[style=ieee,natbib=true,url=false,doi=false,mincitenames=1,maxcitenames=1,maxbibnames=99]{biblatex}
\AtBeginBibliography{\small}
\usepackage{caption}
\usepackage{subcaption}
\usepackage{amsmath}
\usepackage[detect-all]{siunitx}
\usepackage{amssymb}
\usepackage{cleveref}
\usepackage{wrapfig}

\usepackage{tikz}
\usetikzlibrary{arrows,fit} 
\usetikzlibrary{positioning} 
\usetikzlibrary{arrows.meta,calc,shapes} 
\usepackage{aircraftshapes}

\usepackage{pgfplots}
\usepgfplotslibrary{groupplots}
\pgfplotsset{compat=newest}
\pgfplotsset{every axis legend/.append style={legend cell align=left}}
\pgfplotsset{every axis/.append style={
                    title style={font=\small},
                    tick label style={font=\footnotesize}  
                    }}
\pgfplotsset{every axis label/.style={font=\small}}
\definecolor{pastelMagenta}{HTML}{FF48CF}
\definecolor{pastelPurple}{HTML}{8770FE}
\definecolor{pastelBlue}{HTML}{1BA1EA}
\definecolor{pastelSeaGreen}{HTML}{14B57F}
\definecolor{pastelGreen}{HTML}{3EAA0D}
\definecolor{pastelOrange}{HTML}{C38D09}
\definecolor{pastelRed}{HTML}{F5615C}

\title{Risk-Driven Design of Perception Systems}

\author{%
  Anthony L. Corso \thanks{equal contribution} \\
  Department of Aeronautics and Astronautics\\
  Stanford University\\
  Stanford, CA \\
  \texttt{acorso@stanford.edu} \\
  \And
  Sydney M. Katz \footnotemark[1] \\
  Department of Aeronautics and Astronautics\\
  Stanford University\\
  Stanford, CA \\
  \texttt{smkatz@stanford.edu} \\
  \And
  Craig Innes \\
  School of Informatics\\
  University of Edinburgh\\
  Edinburgh, UK \\
  \texttt{craig.innes@ed.ac.uk} \\
  \And
  Xin Du \\
  School of Informatics\\
  University of Edinburgh\\
  Edinburgh, UK \\
  \texttt{x.du@ed.ac.uk} \\
  \And
  Subramanian Ramamoorthy \\
  School of Informatics\\
  University of Edinburgh\\
  Edinburgh, UK \\
  \texttt{s.ramamoorthy@ed.ac.uk} \\
  \And
  Mykel J. Kochenderfer \\
  Department of Aeronautics and Astronautics\\
  Stanford University\\
  Stanford, CA \\
  \texttt{mykel@stanford.edu} \\
}

\begin{document}

\maketitle

\begin{abstract}
Modern autonomous systems rely on perception modules to process complex sensor measurements into state estimates. These estimates are then passed to a controller, which uses them to make safety-critical decisions. It is therefore important that we design perception systems to minimize errors that reduce the overall safety of the system. We develop a risk-driven approach to designing perception systems that accounts for the effect of perceptual errors on the performance of the fully-integrated, closed-loop system. We formulate a risk function to quantify the effect of a given perceptual error on overall safety, and show how we can use it to design safer perception systems by including a risk-dependent term in the loss function and generating training data in risk-sensitive regions. We evaluate our techniques on a realistic vision-based aircraft detect and avoid application and show that risk-driven design reduces collision risk by \SI{37}{\percent} over a baseline system.
\end{abstract}

\section{Introduction}
The design of reliable perception systems is a key challenge in the development of safety-critical autonomous systems~\cite{rosique2019systematic, van2018autonomous}. Modern perception systems are often required to predict state information from complex, high-dimensional inputs such as images or LiDAR data \cite{jung2018perception, li2020lidar, opromolla2021visual}. This information is then passed to a controller, which uses the state estimate to make safety-critical decisions. For example, vision-based perception systems have been proposed for detect and avoid applications in aviation~\cite{opromolla2021visual}.
It is important that perception systems produce accurate estimates, and these systems are typically trained to minimize overall perceptual error using a regression loss on a set of training data. This training method, however, fails to account for the effect of perceptual errors on the performance of the fully-integrated, closed-loop control system. In particular, not all errors will have an equal effect on overall performance. While some errors pose minimal risk, others may lead to catastrophic outcomes \cite{Julian2020, Corso2019itsc}. 

\textbf{Contribution} \ \ In this work, we develop a technique to design perception systems that are sensitive to the overall risk of the closed-loop system in which they operate. We formulate a risk function that quantifies the downstream effect on safety of making a given perceptual error in a given state. We then show how the risk function can be used to design safer perception systems by incorporating it into the loss function during training and using it to focus data collection efforts on the most risk-sensitive regions of the state space. Finally, we analyze the impact of our risk-driven design approach on the safety of a realistic vision-based aircraft detect and avoid system. We show that our approach is able to reduce collision risk by up to \SI{37}{\percent} over a baseline perception system.

\section{Background}\label{sec:background}
This work uses a notion of conditional value at risk (CVaR) and a Markov decision process (MDP) formulation to estimate the effect of perceptual errors on the safety of a closed-loop system. This section details the necessary background on these topics.

\textbf{Conditional Value at Risk (CVaR)} \ \ Let $X$ be a bounded random variable and $F(x) = P(X \leq x)$ be its cumulative distribution function. The value at risk (VaR) is the highest value that $X$ is guaranteed not to exceed with probability $\alpha$ written as
\begin{equation}
    \text{VaR}_\alpha(X) = \min \{x \mid F(x) \geq \alpha\}
\end{equation}
Using the VaR, we can derive the CVaR for a given $\alpha \in (0, 1)$, which represents the expected value of the top $1 - \alpha$ quantile of the probability distribution over $X$. More formally,
\begin{equation}\label{eq:CVaR}
    \text{CVaR}_\alpha(X) = \mathbb{E}[X \mid X \geq \text{VaR}_\alpha(X)]
\end{equation}
If $X$ is random variable that represents cost, $\text{CVaR}_\alpha(X)$ will correspond to the expected value of the worst $\alpha$-fraction of costs. As $\alpha$ approaches \num{1}, $\text{CVaR}_\alpha(X)$ approaches the cost of the worst-case outcome. As $\alpha$ approaches \num{0}, $\text{CVaR}_\alpha(X)$ approaches the expected value of $X$. CVaR has some desirable mathematical properties~\cite{majumdar2020should} that have often made it the preferred risk metric in risk-sensitive domains such as finance \cite{pflug2000some, chow2015risk, keramati2020being} and robotics~\cite{majumdar2020should}.

\textbf{Markov Decision Process (MDP)} \ \ An MDP is a way of encoding a sequential decision making problem in which an agent's action at each time step depends only on its current state \cite{DMU}. An MDP is defined by the tuple $(S, A, T, C, \gamma)$, where $S$ is the state space, $A$ is the action space, $T(s' \mid s, a)$ is the probability of transitioning to state $s'$ given that we are in state $s$ and take action $a$, $C(s,a)$ is the cost of taking action $a$ in state $s$, and $\gamma$ is the discount factor. A stochastic policy for an MDP, written as $\pi(a \mid s)$, denotes the likelihood of taking action $a$ in state $s$ and induces a distribution over future costs. We define a cost distribution function $Z^\pi(s,a)$ that maps a given state $s$ and action $a$ to a random variable representing the sum of discounted future costs obtained by taking action $a$ from state $s$ and subsequently following policy $\pi$ as follows
\begin{equation}
    Z^\pi(s_t,a_t) = C(s_t, a_t) + \sum_{t^\prime = t+1}^\infty \gamma^{t^\prime} C(s_{t^\prime}, a_{t^\prime})
\end{equation}
where $s_{t^\prime+1} \sim T(\cdot \mid s_{t^\prime}, a_{t^\prime})$ and $a_{t^\prime} \sim \pi(\cdot \mid s_{t^\prime})$. To evaluate a given policy $\pi$, we calculate a state-action value function, denoted $Q^\pi(s, a)$, as the expected sum of discounted future costs
\begin{equation} \label{eq:expected_return}
    Q^\pi(s,a) = \mathbb{E}\left[ Z^\pi(s,a) \right]
\end{equation}
In traditional reinforcement learning, $Q^\pi$ is the optimization objective~\cite{DMU}; however, in safety-critical domains we may wish evaluate a policy based on an upper quantile of the worst-case outcomes.
In this case, rather than calculating the expectation of future costs, we define a new state-action value function $Q^\pi_\alpha(s, a)$ that represents the CVaR of the distribution of future costs as
\begin{equation}\label{eq:cvar_return}
Q^\pi_\alpha(s, a) = \text{CVaR}_\alpha[Z^\pi(s, a)]
\end{equation}
Intuitively, $Q^\pi_\alpha(s, a)$ is the expected value of the $\alpha$-percentile of worst-case outcomes. We note that when $\alpha=0$, \cref{eq:cvar_return} reduces to \cref{eq:expected_return}. One way to solve for $Q^\pi_\alpha(s, a)$ is to first solve for the cost distribution function $Z^\pi(s,a)$, which can be done using techniques from distributional reinforcement learning \cite{bdr2022}. The CVaR can then be computed using this distribution. Other techniques that do not require explicitly solving for the distribution over costs may also be applied~\cite{chow2015risk, haskell2015convex}.

\section{Approach}
\begin{figure}[t]
    \centering
    \begin{subfigure}{0.45\textwidth}
        \input{system}
        \caption{System Architecture \label{fig:system}}
    \end{subfigure}
    \begin{subfigure}{0.45\textwidth}
        \input{adversarial}
        \caption{Abstracted Perception MDP \label{fig:adversarial}}
    \end{subfigure}
    \caption{Problem formulation. \label{fig:formulation}}
\end{figure}
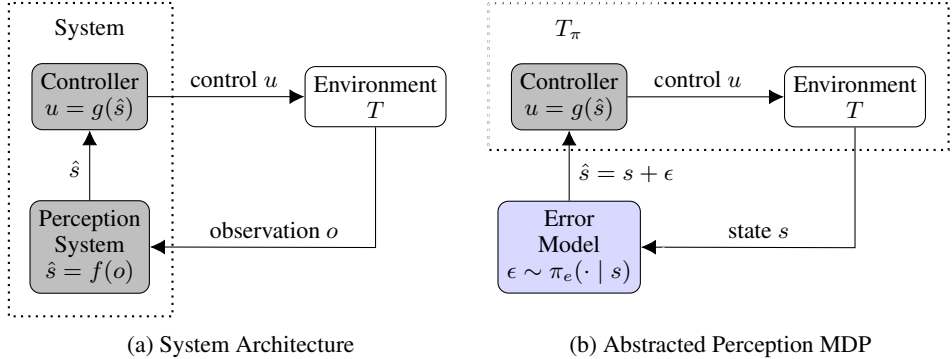
We consider the setting shown in \cref{fig:system}, in which a system is composed of a perception module $f$ and a controller $g$. Given an observation $o \in \mathcal O$ of the true state $s \in \mathcal S$, the perception system produces an estimate of the state $\hat s = f(o)$. The controller produces a control input $u = g(\hat s)$ and the system transitions to state $s'$ with probability $T(s' \mid s, u)$. We assume that the controller $g$ is given, and we wish to design a perception module $f$ such that the overall system satisfies a safety property. 

\begin{figure}[t]
    \centering
    \input{pendulum_overview}
    \caption{Overview of the inverted pendulum example system. \label{fig:pendulum_overview}}
\end{figure}
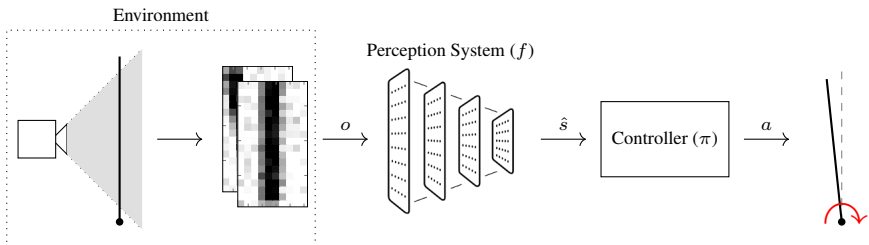

\textbf{Pendulum example} \ \ We will use a vision-based inverted pendulum problem as a running example to describe our approach (see \Cref{fig:pendulum_overview}). We assume that a camera produces noisy image observations of the pendulum's current state $s = [\theta, \omega]$ where $\theta$ is its angle from the vertical and $\omega$ is its angular velocity. To estimate both position and velocity information from the images, we use two consecutive image frames for each observation. The perception system is a multi-layer perceptron (MLP) that takes in the image observations and produces an estimate of the state. Using the state estimate, the controller selects a torque to keep the pendulum upright. We specify a safety property that requires that $|\theta| < \pi/4$ at all time steps. We use a rule-based controller that satisfies the safety property under perfect perception; however, image noise, image downsampling, and the finite capacity of the MLP make perception errors inevitable, and the goal is to design a perception system that limits the occurrence of high risk errors. \Cref{app:pend_controller} summarizes the inverted pendulum model.

\subsection{Overview} 

The key insight in our approach is to abstract the perception system using a notional model of perception errors and analyze the effect of specific perception errors on the performance of the closed-loop system. As shown in 
\cref{fig:adversarial}, we model the errors $\epsilon \in \mathcal E$ of the abstracted perception system as a stochastic policy $\pi_e(\epsilon \mid s)$ in an MDP. We note that this representation implicitly takes into account the possible observations for state $s$ by directly outputting the errors the perception system could make when processing each observation. The cost function of the MDP is chosen to reflect the risk of a being in a state with respect to a safety property. By formulating the problem in this way, we can use the policy evaluation techniques outlined in \cref{sec:background} to evaluate the risk of making perception error $\epsilon$ in state $s$.
Given that $\pi_e$ is an abstraction of the perception system, it will not exactly reflect the true distribution of perception errors. To account for this mismatch, we are conservative in our risk estimate by evaluating the policy according to \cref{eq:cvar_return}. Once computed, the risk function represents the downstream impact of making a given perception error in the current state. This function can therefore be used in a supervised learning setting to encourage a model to avoid making high-risk errors. We also show that the risk function can be used to identify error-sensitive regions of the state space, from which additional training data can be collected.

\subsection{Risk Function} \label{sec:risk_func}
Our goal is to define a risk function $\rho(s, \epsilon)$ that quantifies the risk of making perception error $\epsilon$ in state $s$ when using a controller $g$. We model the perception error as a stochastic policy $\pi_e(\epsilon \mid s)$ that reflects a notional distribution over errors in state $s$. The controller produces an action based on the perceived state $\hat s$, which is computed from $s$ and $\epsilon$. For additive noise, the resulting transition function for the abstracted perception MDP is
\begin{equation}
    T_\pi(s' \mid s, \epsilon) = T(s^\prime \mid s, g(s + \epsilon))
\end{equation}

The error model $\pi_e$ can be designed to match the performance of a previously trained baseline perception system or using domain knowledge. If we are confident that our choice closely matches the errors we will see when designing the perception system, we can set our risk function to match the state action-value function in \cref{eq:expected_return}. However, if we have not yet designed the perception system, there will be a mismatch between our estimate of $\pi_e$ and the true distribution of perception errors.

In these scenarios, we may want a more conservative risk function in which we consider the CVaR of the distribution over future costs when making perception error $\epsilon$ in state $s$ given that all future perception errors are distributed according to $\pi_e$. Therefore, we select our risk function as follows
\begin{equation}
    \rho^{\pi_e}_\alpha(s, \epsilon) = Q^{\pi_e}_\alpha(s, \epsilon)
\end{equation}
where $Q^{\pi_e}_\alpha(s, \epsilon)$ refers to the CVaR state-action value function defined in \cref{eq:cvar_return} for the abstracted perception MDP. The parameter $\alpha$ smoothly controls the degree of conservatism. When $\alpha = 0$, the risk function is equivalent to the expected cost when following the notional perception model. When $\alpha=1$, we consider only the worst-case sequence of possible perception errors.

To solve for the CVaR state-action value function we apply distributional dynamic programming to compute the distribution over costs. 
Distributional dynamic programming (alg. 5.3 in \citet{bdr2022}) resembles value iteration where the Bellman operator is replaced with a distributional version to update a parameterized distribution. In this work, we approximate the distribution over costs using a categorical distribution with a fixed number of discrete costs. Additionally, we discretize the states and actions and use a linear weighting function for local approximation. To extend our method to problems with larger state and action spaces, we could apply function approximation techniques developed in prior work~ \cite{bellemare2017distributional, dabney2018distributional, dabney2018implicit}, but we leave this as future work. Once we solve for the cost distributions, we can calculate $\rho^{\pi_e}_\alpha(s, \epsilon)$ for any value of $\alpha$ using \cref{eq:CVaR} with minimal additional computation.

\begin{figure}[t]
    \centering
    \input{pend_cvar_slices_alpha}
    \caption{Values of $\rho^{\pi_e}_\alpha([0.2, 0.0], \epsilon)$ for different values of $\alpha$. The state and an example error are depicted on the left. \label{fig:pend_cvar_slices}}
\end{figure}
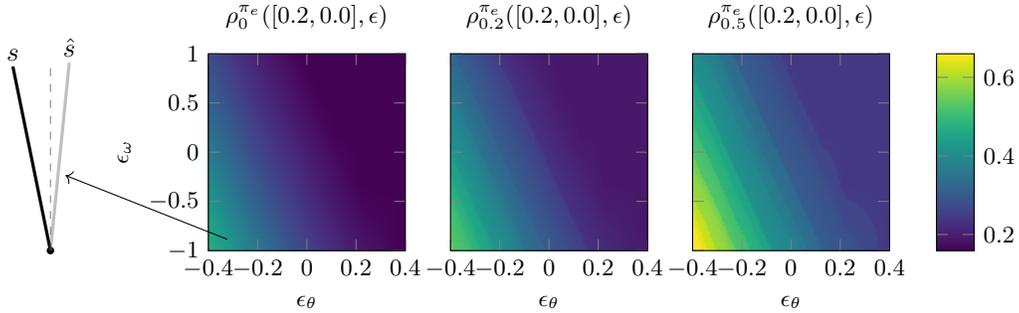

\textbf{Pendulum example} \ \ For the inverted pendulum system, we model the perception errors $\epsilon = [\epsilon_\theta, \epsilon_\omega]$ as additive noise on the true state ($\hat{s} = s + \epsilon$) distributed according to a multivariate Gaussian distribution with zero mean and diagonal covariance. The cost function penalizes the absolute deviation from the vertical $|\theta|$. See \cref{app:pend_risk_func} for more detail.
\Cref{fig:pend_cvar_slices} shows the values of the pendulum risk function at the state $s = [0.2, 0.0]$ over the set of errors for multiple values of $\alpha$. A higher value of $\alpha$ will put more weight on worst-case perception errors in the future, so as $\alpha$ increases, the overall risk values increase. The risk function also provides a notion of \textit{which} perception errors are most risky in a given state. For example, the optimal action when $s = [0.2, 0.0]$ is to apply a negative torque to move the pendulum closer to upright, but if perception errors are negative, the resulting $\hat s$ may indicate that the pendulum is at a smaller angle and already moving in the negative direction as shown in the leftmost image on \cref{fig:pend_cvar_slices}. This estimate may cause the controller to produce a smaller torque or even a torque in the positive direction, leading to high risk of falling over.


\subsection{Risk-Sensitive Loss Function}\label{sec:risk_loss}
Perceptual errors are inevitable whenever there is sensor noise, occlusions, limited training data, or limited model capacity. Our goal is to ensure that the errors that are made by the perception system are in the direction of lower, rather than higher, risk.
We incorporate this notion into our design by formulating a risk-sensitive loss function. Let $\mathcal L(s, \hat s)$ be the standard loss function for the baseline perception system such as mean squared error for regression tasks or cross entropy for classification tasks. We define a risk-sensitive loss function for a single datapoint with additive noise as follows
\begin{equation}\label{eq:risk_loss}
    \mathcal L_R(s, \hat s) = \mathcal L(s, \hat s) + \lambda \rho^{\pi_e}_\alpha (s, \hat s - s)
\end{equation}
where $\lambda$ is a hyperparameter that controls the relative weighting of each term. The first term in the loss function pushes all perception errors toward zero, while the second term seeks to minimize the risk of the current errors that the model makes. We note that this is different from simply minimizing perception errors in states with high risk values. If the risk function outputs the same value for a given state regardless of perception error, the gradient of the risk term in the loss function will be zero.

\textbf{Pendulum example} \ \ For the inverted pendulum problem, we use a mean squared error baseline loss function, which penalizes positive and negative errors equally. However, a positive and negative error of the same magnitude do not necessarily have an equal effect on the overall risk of the closed-loop system. For example, consider the results shown in \cref{fig:pend_cvar_slices} where the pendulum is positioned at an angle of \num{0.2} from the vertical. The risk function shows that a perception error of $\epsilon_\theta = 0.2$ ($\hat\theta = 0.4$) is less risky than a perception error of $\epsilon_\theta = -0.2$ ($\hat\theta = 0.0$). Intuitively, predicting that the pendulum is tipped further in the positive direction will still result in a negative torque to move the pendulum closer to upright. In contrast, predicting that the pendulum is perfectly upright will cause the controller to output zero torque. The second term in \cref{eq:risk_loss} accounts for this effect.

\subsection{Risk-Driven Data Generation}\label{sec:risk_data}
In addition to considering the directionality of the errors, we can use the risk function we formulated in \cref{sec:risk_func} to identify states that are especially sensitive to perception errors and generate more training data in those regions. In effect, this process gives higher weight to risky states during training. Since training with a regression loss will push all errors toward zero, we define risky states as states in which making a nonzero perception error will result in a high risk value compared to the risk of zero error. We use a weighting function $w_\alpha(s)$ to capture this notion as follows
\begin{equation}\label{eq:risk_weight}
    w_\alpha(s) = \max_{\epsilon \in \mathcal E} \rho^{\pi_e}_\alpha(s, \epsilon) - \rho^{\pi_e}_\alpha(s, 0)
\end{equation}
We note that $w_\alpha(s)$ is not the same as (and not necessarily correlated with) the risk of being in state $s$. Instead, $w_\alpha(s)$ represents the risk of making a nonzero perception error in state $s$. For instance, states in which failure is inevitable regardless of perception error will have high risk values but low weight assigned to them according to $w_\alpha(s)$. To sample data, we use rejection sampling with $w_\alpha(s)$ as the likelihood function.

\begin{figure}[t]
    \centering
    \input{pend_rw}
    \caption{Illustration of \cref{eq:risk_weight}. \label{fig:pend_risk_weights}}
\end{figure}
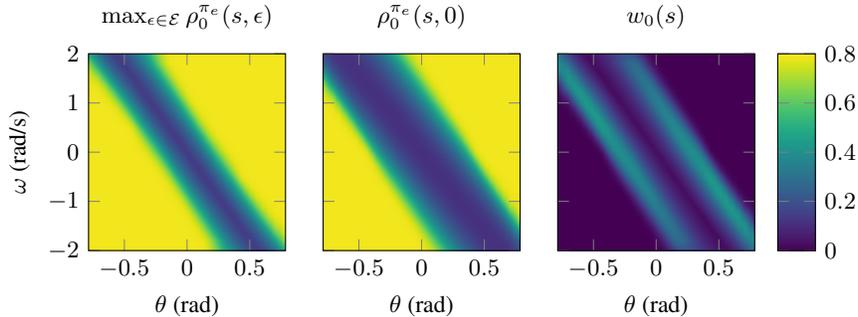

\textbf{Pendulum example} \ \ \Cref{fig:pend_risk_weights} illustrates each term in \cref{eq:risk_weight} for the inverted pendulum example problem. The lower left and upper right regions of the state space represent states in which the pendulum is at a high angle and moving in a direction away from the vertical. Due to underactuation of the pendulum system, the states at these extremes are guaranteed to lead to failure regardless of perception error. States in the central band of each plot, on the other hand, constitute states in which the pendulum is likely to remain upright as long as proper control is applied. This band widens when we consider the effect of having zero perception error compared to the worst-case perception error. By subtracting these quantities, we can understand in which states having zero perception error is most critical. Given a limited data budget, we can focus our data collection on these states by sampling data according to $w_\alpha(s)$.

\subsection{Hypotheses}
We design the experiments in this work to test the following two hypotheses:
\begin{itemize}
    \item \textbf{H1}: A perception module trained using the risk-sensitive loss function proposed in \cref{sec:risk_loss} will result in a safer closed-loop system than one trained using the baseline loss on the same dataset.
    \item \textbf{H2}: A perception system trained on data sampled according to the risk-driven data generation technique proposed in \cref{sec:risk_data} will result in a safer closed-loop system than one trained on data uniformly sampled throughout the state space.
    
\end{itemize}

\textbf{Pendulum example} \ \ We evaluate the inverted pendulum  by measuring the mean time to failure over \num{100} trajectories each with \num{500} time steps from random initial states. We report the mean and standard error of this metric over \num{5} trials with \num{500} being the maximum attainable value. To test \textbf{H1}, we train a baseline perception system using a mean squared error loss function and a risk-driven perception system using the loss function described in \cref{sec:risk_loss} on \num{10000} uniformly sampled data points. To make the perception problem more challenging, we add noise to the images. The risk sensitive loss function improved the mean time to failure from the baseline of $234 \pm 62$ to $492 \pm 6$. To test \textbf{H2} on the inverted pendulum system, we limit the perception system to a data budget of \num{50} training examples. We train a baseline perception system on data sampled uniformly throughout the state space and a risk-driven perception system on data sampled according to the weighting function shown in \cref{fig:pend_risk_weights}. The mean time to failure for the baseline system is $152 \pm 89$, while the time to failure for the system trained of the risk-driven data was \num{500} for all trials. See \cref{app:pend_results} for full results.


\section{Vision-Based Detect and Avoid Application}
Aircraft collision avoidance systems use information gathered from sensors to detect intruding aircraft and issue advisories for safe collision avoidance maneuvers \cite{kochenderfer2012next}. Traditional sensors used for surveillance and tracking include ADS-B, onboard radar, and transponders \cite{owen2019acas}; however, autonomous aircraft require additional sensors both for redundancy and to replace the visual acquisition typically performed by the pilot. For this reason, vision-based traffic detection systems have been proposed, in which intruding aircraft are detected from images taken by a camera sensor mounted on the aircraft \cite{opromolla2021visual, james2018learning}. In this section, we will apply our risk-driven design techniques to improve the safety of a vision-based detect and avoid (DAA) system. The code for this work can be found at \href{https://github.com/sisl/RiskDrivenPerception}{https://github.com/sisl/RiskDrivenPerception}.

\begin{figure}[htb]
    \centering
    \input{daa_overview}
    \caption{Overview of the vision-based detect and avoid system application. \label{fig:daa_overview}}
\end{figure}
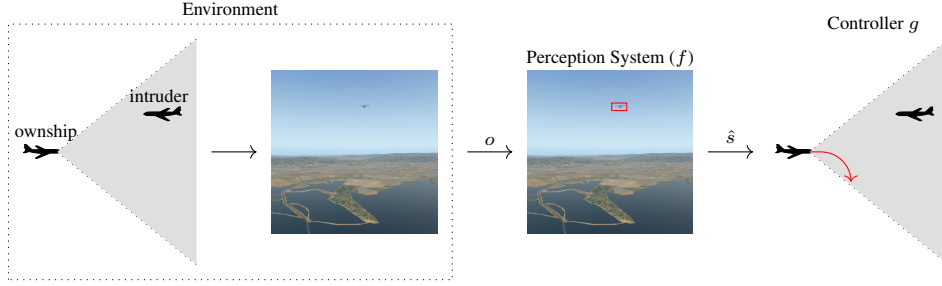

\subsection{Problem Setup}
\Cref{fig:daa_overview} outlines the components of a vision-based DAA system. A camera mounted on the aircraft (which we refer to as the ownship) produces image observations of its surroundings. These images are then passed through a perception system to produce an estimate of the state for the controller. The collision avoidance controller $g$, which is loosely based on the ACAS X family of collision avoidance systems \cite{kochenderfer2012next}, commands vertical maneuvers based on an estimate of the relative position and velocity of an intruder aircraft. Specifically, the state is defined as $s = [p, h, \dot h, a_\text{prev}, \tau]$ where $p$ is a binary variable describing whether an intruder is present, $h$ is the relative altitude of the intruder, $\dot h$ is the relative vertical rate of the intruder, $a_\text{prev}$ is the previous action, and $\tau$ is the time to loss of horizontal separation. The controller selects an action $u \in \{\textsc{coc}, \textsc{climb}, \textsc{descend}\}$ where \textsc{coc} represents clear of conflict. If $p=0$, the action is always \textsc{coc}.

We assume that the state estimate for the controller is produced using two steps. First, the image is passed through a perception network that is responsible for detecting other aircraft (which we refer to as intruders) and producing a bounding box. Once an intruder has been detected, the ownship interrogates it to determine the rest of the state. We focus our risk-driven design efforts on the detection component of the perception system. We train a baseline network using the YOLOv5 algorithm \cite{Redmon_2016_CVPR, glenn_jocher_2022_6222936} on \num{10000} simulated images in which the intruder location is sampled uniformly within the ownship field of view. For additional details on the controller, network architecture, computational resources, and training, see \cref{app:daa}.


\subsection{Risk-Driven Design}
An object detector can make two types of errors: false positives and false negatives. While false positives have an effect on the efficiency of the system, they do not affect the risk of a collision. We therefore focus on quantifying the risk of of false negatives in the presence of an intruder. We define the perception error $\epsilon$ as zero when an intruder is properly detected and one when it is missed, leading to a perceived state
\begin{equation}
\hat{s} = [1 - \epsilon, h, \dot h, a_\text{prev}, \tau]
\end{equation}
The notional error model $\pi_e(\epsilon \mid s)$ is a Bernoulli distribution with a success probability based on the detection performance of the baseline model (see \cref{app:daa} for more detail). To solve for $Q^{\pi_e}_\alpha(s, \epsilon)$, we discretize the state space and apply approximate dynamic programming.


\textbf{Risk-sensitive loss function} \ \ In addition to outputting the bounding box coordinates, the detection network outputs an objectness score $\hat p$ representing the model's confidence that an aircraft is present in the frame.
\begin{wrapfigure}[]{r}{0.4\textwidth}
    \centering
    \input{daa_rw}
    \caption{DAA weighting function. \label{fig:daa_rw}}
\end{wrapfigure}
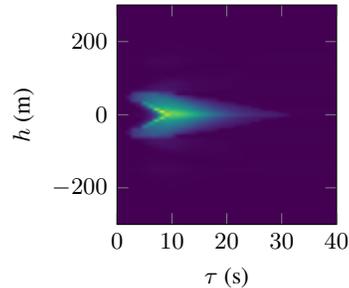
The model detects an aircraft after checking if $\hat p$ exceeds a threshold;
however, this check poses a numerical challenge when differentiating 
the risk sensitive loss since the gradient is undefined at the decision threshold and is zero everywhere else. 
To avoid this problem we evaluate the risk of $\hat p$ directly by interpolating between the risk of not detecting and detecting the intruder
\begin{equation}\label{eq:daa_risk_func}
    \rho_\alpha^{\pi_e}(s, \hat p) = (\hat p) Q_\alpha^{\pi_e}(s, 0) + (1 - \hat p) Q_\alpha^{\pi_e}(s, 1)
\end{equation}
The gradient of risk function in \cref{eq:daa_risk_func} with respect to $\epsilon$ is equal to the risk weight, so the model will be incentivized to be more confident in states where a detection has a significant effect on overall risk.

\textbf{Risk-driven data generation} \ \ \Cref{fig:daa_rw} shows the weighting function when $\alpha=0$ and an intruder is present. States with high weights correspond to states where not detecting the intruder is significantly more risky than detecting the intruder. While a collision is imminent in states with low values of $\tau$ and $|h|$, these states have low weights since a collision will occur regardless of whether the intruder is detected. The states with the highest risk weights are states in which there is just enough time to resolve an impending collision as long as the intruder is detected.

\subsection{Evaluation}
The safety of aircraft collision avoidance systems is often assessed using Monte Carlo analysis on airspace encounter models \cite{TCASMonteCarlo,ACASXMonteCarlo}. Encounter models are probabilistic representations of typical aircraft behavior during a close encounter with another aircraft. To analyze the safety of a particular collision avoidance system, we can simulate the system on a set of encounters and count the number of times a near mid-air collision (NMAC) occurs. We define an NMAC as a simultaneous loss of aircraft separation to within \SI{50}{\meter} vertically and \SI{100}{\meter} horizontally. For the evaluation in this work, we use model consisting of \num{1000} pairwise encounters in which the ownship and intruder follow straight line trajectories with various relative geometries. All encounters result in an NMAC if no collision avoidance actions are taken (see \cref{app:enc_model}).

\begin{figure}[t]
    \centering
    \input{daa_results_v2}
    \caption{Number of NMACs and cumulative risk for each perception system. \label{fig:daa_results}}
\end{figure}
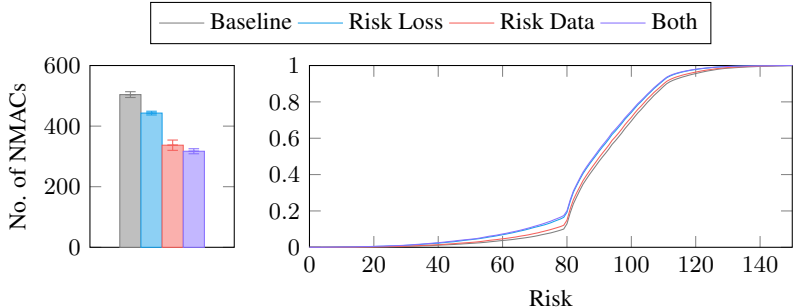

We trained perception networks using four perception design techniques: no risk awareness (baseline), risk-sensitive loss function, risk-driven data generation, and both a risk-sensitive loss function and risk-driven data generation. We performed three trials for each perception design method. The first section of \cref{fig:daa_results} shows the number of encounters that result in an NMAC for each technique. In support of \textbf{H1}, training with the risk-sensitive loss function (blue) results in a slight increase in safety. In support of \textbf{H2}, training on data generated using the weighting function (red) results in a \SI{33}{\percent} decrease in the number of NMACs over the baseline.  With a \SI{37}{\percent} decrease in the number of NMACs compared to the baseline, the perception systems that were developed using both risk-driven techniques (purple) are slightly safer than using either technique individually.
\Cref{fig:daa_results} also shows the cumulative distribution over the risks of the errors made by each perception system on the encounter set. This result implies that even small shifts in overall risk can have significant impacts on safety.

\begin{figure}[t]
    \centering
    \input{daa_example}
    \caption{Example encounter that was resolved by the risk-sensitive systems but resulted in an NMAC for the baseline system. The red line denotes the intruder trajectory, while the gray and purple lines denote the ownship trajectory when using the baseline and risk-sensitive perception systems respectively. Left: overhead view of the encounter. Bottom right: altitude of each aircraft over time along with the risk of each perception error. The dotted markings indicate time steps in which the intruder was detected. Top right: perception outputs of each system at four snapshots in time. \label{fig:daa_ex}}
\end{figure}
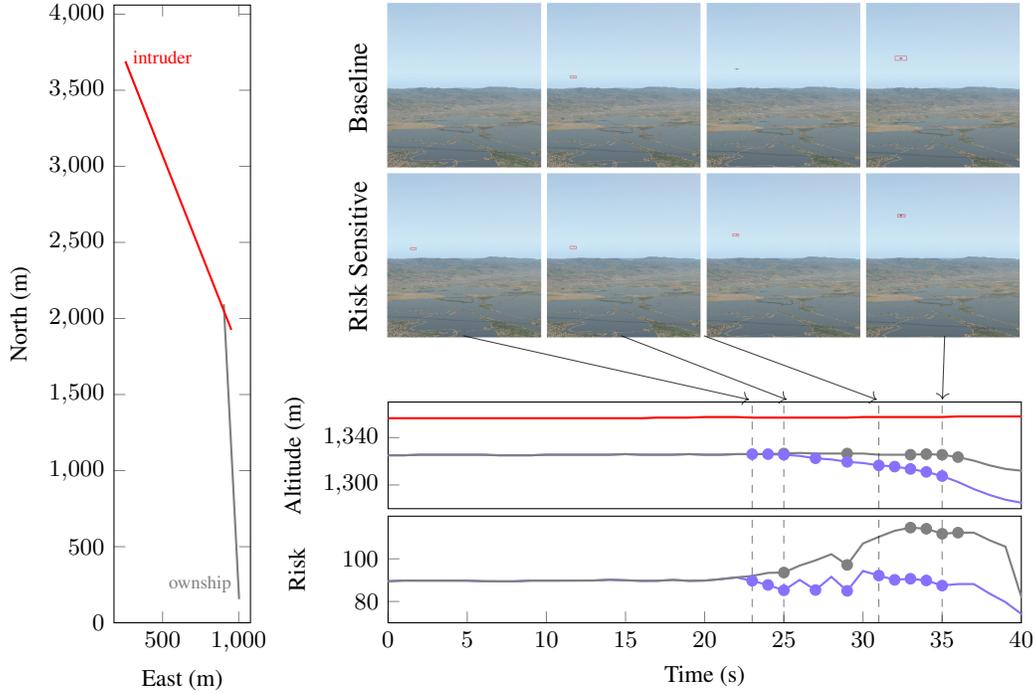

\Cref{fig:daa_ex} shows an example encounter that resulted in NMAC for the baseline system but was resolved by the risk-sensitive systems. The risk-sensitive system (purple) detects the intruder aircraft sooner and more often than the baseline. Moreover, the plot at the bottom of the figure shows that the risks of the perceptual errors made by the baseline system are much higher than the risks of the errors made by the risk-sensitive system.

\section{Related Work}
We discuss prior work in perception system evaluation, risk-aware control, and task-aware prediction.

\textbf{Perception system evaluation} \ \ The high-dimensional, complex nature of perception inputs makes closed-loop safety evaluation of perception systems especially challenging. This challenge was first noted in early work on computer vision, in which systems were evaluated by propagating uncertainty in a set of input images through each computer vision component \cite{liu2005use, haralick1992performance}. Similarly, a number of recent works have focused on finding adversarial perturbations that will cause misclassification of single input images \cite{machado2021adversarial}. 
\citet{Julian2020}, in contrast, use adaptive stress testing (AST) \cite{Corso2021survey} to find sequences of adversarial perturbations that will lead a vision-based aircraft taxi system to failure rather than focusing on single images in isolation. However, this approach is only able to find failures and cannot guarantee safety if no failures are found. \citet{Katz2021dasc} provide approximate formal guarantees of closed-loop safety properties by using a generative model to approximate the set of plausible inputs to a perception system. \citet{dreossi2019compositional} address the complexity of analyzing the closed-loop system by analyzing the controller and perception system separately before combining the results to find failures in the overall system. This work draws inspiration from this decoupled approach.

\textbf{Risk-aware control} \ \ In safety-critical domains, it is desirable to design controllers that perform well even in worst-case conditions. This problem has been addressed through safety-constrained MDPs \cite{wachi2018safe} and shielding techniques in reinforcement learning \cite{alshiekh2018safe}.
Other risk-aware control approaches use a notion of CVaR to design controllers. CVaR MDPs, for instance, have been used to produce control policies that are robust to modeling errors and worst-case outcomes \cite{chow2015risk, haskell2015convex}. Distributional reinforcement learning approaches estimate the distribution of returns~\cite{bellemare2017distributional}.  Using the estimate of the return distribution, it is possible to compute risk-related metrics such as CVaR and produce risk-aware reinforcement learning policies \cite{chow2014algorithms, dabney2018distributional, dabney2018implicit, keramati2020being}. In this work, rather than using a notion of risk to design control policies, we use it to design safe perception systems.

\textbf{Task-aware prediction} \ \ Previous work has explored the benefit of using task-aware metrics to inform the design of predictive models. 
For example, \citet{lambert2020objective} note that more accurate dynamics models in reinforcement learning do not necessarily correlate with higher rewards, and \citet{bansal2017goal} show improved performance when learning a dynamics model that achieves the best control rather than learning the most accurate model. Multiple techniques have been proposed to augment traditional training loss functions with task-specific metrics to design models that are not only accurate but also satisfy domain constraints \cite{innes2020elaborating, pmlr-v97-fischer19a, leung2020back}.
In the context of trajectory prediction, \citet{mcallister2022control} propose control-aware prediction objectives, which take into account the downstream effects of the predictions on controller performance. These works motivate the need for task-aware objectives but are not used with image-based perception. In the context of perception, \citet{greiffenhagen2001design} introduce the idea of connecting perception errors to overall system performance for the design and evaluation of a system that uses traditional computer vision \cite{greiffenhagen2000statistical, greiffenhagen2001design, greiffenhagen2001systematic}. Furthermore, \citet{philion2020learning} develop planner-centric metrics for an machine-learning based object detection system and show their benefit over traditional metrics used to evaluate object detectors. However, they focus only on the evaluation of perception systems, while this work focuses on their design.

\section{Conclusion}
\label{sec:conclusion}
In this work, we presented a methodology for risk-driven design of safety-critical perception systems. We formulated a risk function that measures the effect of perception errors on the closed-loop performance of the fully-integrated perception and control system. We then showed how to use that risk function during the design process by incorporating it into the loss function and developing a risk-driven data generation technique. We demonstrated our approach on a realistic vision-based aircraft detect and avoid system and showed that our techniques could increase safety by \SI{37}{\percent} over a baseline system. We note that while the methods presented here focus on designing safe perception systems, they do not represent safety guarantees. Therefore, perception systems designed in this manner should still be put through additional testing and safety validation before deployment. A limitation of this work is that we assume that the perception system is Markov, which limits the applicability of our technique to perception systems that use filtering to produce state estimates. Future work will address this limitation.


\begin{ack}
The NASA University Leadership Initiative (grant \#80NSSC20M0163) provided funds to assist the authors with their research. This research was also supported by the National Science Foundation Graduate Research Fellowship under Grant No. DGE–1656518. Any opinion, findings, and conclusions or recommendations expressed in this material are those of the authors and do not necessarily reflect the views of any NASA entity or the National Science Foundation. This work was also supported by a grant from the UKRI Strategic Priorities Fund to the UKRI Research Node on Trustworthy Autonomous Systems Governance and Regulation (EP/V026607/1, 2020-2024).
\end{ack}







\typeout{}
\printbibliography

\clearpage
\appendix

\section{Inverted Pendulum Details}\label{app:inv_pend}
\subsection{Inverted Pendulum Model and Controller}\label{app:pend_controller}
The inverted pendulum system is modeled using state $s = [\theta, \omega]$ where $\theta$ is the angle of the pendulum from the vertical, and $\omega$ is its angular velocity. The discrete-time dynamics for the inverted pendulum are represented as follows
\begin{equation}
    \begin{split}
        \theta_{t+1} & = \theta_t + \omega_t \Delta t \\
        \omega_{t+1} & = \omega_t - \frac{3g}{2\ell}\sin(\theta_t + \pi) + \frac{3a}{m\ell^2}\Delta t
    \end{split}
\end{equation}
where $g$ is the acceleration due to gravity, $\ell$ is the length of the pendulum, $m$ is the mass of the pendulum, $\Delta t$ is the time step, and $a$ is the input torque from the controller. In this work, we use 
$g = \SI{10}{\meter\per\second}$, $\ell = \SI{1}{\meter}$, $m = \SI{1}{\kilogram}$, and $\Delta t = \SI{0.05}{\second}$. We clip $\omega$ such that the magnitude of the angular velocity does not exceed \SI{8}{\radian\per\second}, and we clip the control inputs so that the maximum torque magnitude does not exceed \SI[inter-unit-product =\ensuremath{\cdot}]{2}{\newton\meter}.

We derive a simple rule-based policy to balance the pendulum according to the following two equations
\begin{equation}
    \begin{split}
        \omega_\text{target} & = \text{sign}(\theta) \sqrt{60(1 - \cos(\theta)} \\
        a & = -2\omega + (\omega - \omega_\text{target})
    \end{split}
\end{equation}
where the first equation determines the angular velocity required to move the pendulum from its current angle to an angle of zero. The second equation performs proportional control using this quantity and the current angular velocity. This controller is able to keep the pendulum upright under perfect perception.

\subsection{Inverted Pendulum Risk Function}\label{app:pend_risk_func}
\textbf{Error model} \ \ We model the perception errors $\epsilon = [\epsilon_\theta, \epsilon_\omega]$ as additive noise distributed according to a multivariate Gaussian distribution with zero mean and diagonal covariance. Noting that predicting $\omega$ from successive frames tends to be more difficult for a perception network than prediction $\theta$ according to our baseline model, we set the covariance as
\begin{equation}
    \Sigma = \begin{bmatrix}
                0.2 & 0.0\\
                0.0 & 0.5
              \end{bmatrix}
\end{equation}
To solve for the risk function using distributional dynamics programming, we discretize the error model and approximate it as a categorical distribution. We select \num{11} discrete points for $\epsilon_\theta$ on a logarithmic scale between \num{-0.4} and \num{0.4} (two standard deviations) such that the points are more highly concentrated around zero error. We determine their corresponding weights by computing their corresponding probability densities from the continuous Gaussian model and normalizing.

\textbf{State discretization} We discretize $\theta \in [-\pi/4, \pi/4]$ and $\omega \in [-2, 2]$ using \num{41} points with a higher density of points around zero. To solve for the risk function, we assume an episodic setting in which each episode has \num{20} time steps. To reflect the episodic nature in the solving process, we add a time component to the state such that $s = [t, \theta, \omega]$ that specifies the number of steps left in the episode. For the next state in the dynamics, $t$ is decremented by one.

\textbf{Cost function} \ \ The cost function for the inverted pendulum problem is defined as the absolute angle of the pendulum at the last time step in each episode:
\begin{equation}
    c(s, \epsilon) = \begin{cases}
    |\theta|, & \text{if } t = 0 \\
    0, & \text{if } t > 0
    \end{cases}
\end{equation}
Since we specify that the pendulum fails if $|\theta| > \pi/4$, the cost is bounded between $0$ and $\pi/4$. We select \num{50} discrete cost points on a logarithmic scale within this range for the distributional dynamic programming.

Solving for the cost distributions using dynamic programming takes under a minute on a single Intel Core i7 processor operating at \SI{4.20}{\giga\hertz}. We save the distributions and use linear interpolation and \cref{eq:cvar_return} to evaluate the risk at an arbitrary $\epsilon$. Since the autodiff package we use for risk-sensitive training does not have a straightforward way to differentiate through an interpolation of tabular data, we train a neural network surrogate model of the risk function to use in the risk-sensitive loss. We train a small MLP network with one hidden layer containing \num{64} hidden units for each value of $\alpha$.

\textbf{Perception system training} \ \ All perception systems trained for the pendulum problem were neural networks with two hidden layers with ReLU activations and \num{64} units each. The final layer uses a hyperbolic tangent activation (the labels are scaled to range between \num{-1} and \num{1}). We train using the ADAM optimizer with a learning rate of \num{1e-3}. We train for \num{200} epochs for the risk data experiment and \num{400} epochs for the risk loss experiment.

\subsection{Inverted Pendulum Results}\label{app:pend_results}
\Cref{tab:pend_res} shows the full results for the inverted pendulum experiments. We ran five trials for each experiment, and we report the mean and standard error of the time to failure for each experiment. Since we simulate \num{100} episodes of length \num{500} for each evaluation the highest possible mean time to failure is \num{500}, which would indicate that the pendulum stayed upright for all time steps. The risk-sensitive design indicates a benefit for most values of $\alpha$; however, we note that performance seems to degrade when $\alpha$ becomes too large, which is likely due to being overly pessimistic. If we always assume the worst-case outcomes, there may not be any states in which a failure can be prevented. In this case, the pendulum system will not be incentivized to produce accurate estimates at any state.
\begin{table}[htb]
    \centering
    \caption{Mean time to failure for pendulum perception systems. \label{tab:pend_res}}
    \begin{tabular}{@{}lll@{}}
          \toprule
          & Risk Loss & Risk Data \\
          \midrule
          Baseline & $234 \pm 62$ & $152 \pm 89$ \\
          \arrayrulecolor{black!30}\midrule
          Risk Sensitive ($\alpha=0.00$) & $\mathbf{492 \pm 6}$ & $334 \pm 102$ \\
          Risk Sensitive ($\alpha=0.20$) & $473 \pm 18$ & $\mathbf{500 \pm 0}$ \\
          Risk Sensitive ($\alpha=0.50$) & $486 \pm 12$ & $404 \pm 86$ \\
          Risk Sensitive ($\alpha=0.80$) & $489 \pm 8$ & $438 \pm 59$ \\
          Risk Sensitive ($\alpha=0.99$) & $391 \pm 59$ & $273 \pm 65$ \\
          \arrayrulecolor{black}\bottomrule
    \end{tabular}
\end{table}

\section{Vision-Based Detect and Avoid Details}\label{app:daa}
\subsection{Collision Avoidance Controller}
The collision avoidance controller takes in a state $s = [p, h, \dot h, a_{\text{prev}}, \tau]$ where $p$ is a binary variable describing whether an intruder is present, $h$ is the relative altitude of the intruder, $\dot h$ is the relative vertical rate of the intruder, $a_\text{prev}$ is the previous action, and $\tau$ is the time to loss of horizontal separation. The relative states, $h$ and $\dot h$, defined to be the ownship quantity minus the intruder quantity. For example, if the intruder is below the ownship, $h$ will be positive. The previous action is contained within the state to penalize undesirable operational characteristics such as reversals in collision avoidance advisory. Finally, the variable $\tau$ is used to summarize the horizontal evolution of the encounter in a single variable. The goal is to have sufficient vertical separation $h$ as $\tau$ gets close to zero to avoid a near mid-air collision (NMAC).

The controller outputs an action $u \in \{\textsc{coc}, \textsc{climb}, \textsc{descend}\}$ where \textsc{coc} represents clear of conflict and the \textsc{climb} and \textsc{descend} actions commands vertical rates. Specifically, the \textsc{climb} action commands a vertical rate of \SI{8}{\meter\per\second}, while the \textsc{descend} action commands a vertical rate of \SI{-8}{\meter\per\second}. For \textsc{coc}, we let $u=0$. Given a state $s_t$ and action $u_t$, we calculate the next state as follows
\begin{equation}
    \begin{split}
        p_{t + 1} & = p_t \\
        h_{t + 1} & = h_t + \dot h_t \\
        \dot h_{t+1} & = \dot h_t + u_t + w \\
        a_{\text{prev}, t+1} & = u_t \\
        \tau_{t+1} & = \tau_t - 1
    \end{split}
\end{equation}
where $w$ represents potential noise in vertical rate due to factors such as pilot response delays. We set this noise to a categorical distribution in which $w=0$ with \num{0.8} probability and $w = \pm 0.5$ with probability \num{0.1} each. We also provide a maximum acceleration to the model and clip $u_t$ based on the current vertical rate to comply with this maximum acceleration.

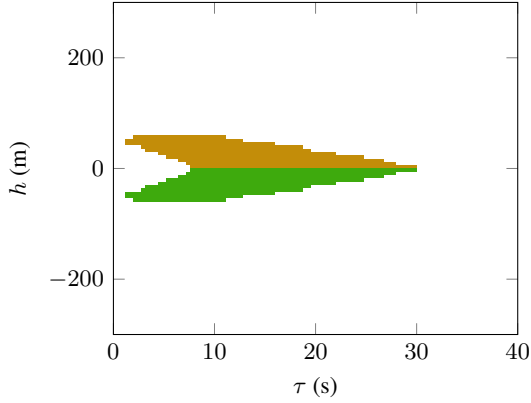
\begin{figure}
    \centering
    \input{daa_pol}
    \caption{Policy for the collision avoidance controller for a slice of the state space where $\dot h = 0$ and $a_\text{prev}=\textsc{coc}$. Orange regions indicate regions of the state space where a \textsc{climb} advisory is issued, green regions indicate \textsc{descend}, and white areas are \textsc{coc}. \label{fig:daa_pol}}
\end{figure}

We develop a controller that is loosely inspired by the ACAS X family of collision avoidance problems by formulating the control problem as an MDP and solving for the optimal policy using dynamic programming \cite{kochenderfer2012next}. The solving process takes under a minute on a single Intel Core i7 processor operating at \SI{4.20}{\giga\hertz}. \Cref{fig:daa_pol} shows the resulting policy over a slice of the state space. As expected, the alerting region shrinks with increasing $\tau$ since there is more time to resolve collisions. No action is taken in the region near $h=0$ and $\tau=0$. This region represents states in which a collision is imminent and taking any action to avoid collision would be futile.

\subsection{Baseline DAA Perception Network}
To train a baseline detection network, we gathered images of intruder aircraft labeled with their corresponding bounding boxes using the X-Plane 11 flight simulator, which has been used in previous work to gather data for a vision-based aircraft taxi scenario \cite{katz2021dataset}. For each data point, we first sample a random ownship position and orientation in the airspace around the Palo Alto airport (PAO). We then sample an intruder position and orientation uniformly within the field of view and record the resulting image. Finally, we compute the corresponding bounding box label based on the relative position of the intruder aircraft. We train the detection networks using the YOLOv5 algorithm \cite{Redmon_2016_CVPR, glenn_jocher_2022_6222936} based on the open source (GNU general public license) PyTorch implementation at \href{https://github.com/ultralytics/yolov5}{https://github.com/ultralytics/yolov5}. For each experiment, we use the YOLOv5s model architecture, which has \num{7.2} million trainable parameters and train on a dataset of \num{10000} images for \num{200} epochs with a batch size of \num{16} using the default hyperparameters. A single training run takes $\approx 11$ hours on a single Intel Core i7 processor operating at \SI{4.20}{\giga\hertz} and adding the risk-sensitive loss component did not significantly increase the required training time.

\subsection{DAA Risk Function}
\textbf{Error model} \ \ We model the perception error for the DAA model as $\epsilon=0$ when the intruder aircraft is detected and $\epsilon=1$ when it is not detected. The notional error model $\pi_e(\epsilon \mid s)$ is a Bernoulli distribution with a success probability corresponding to the probability of detecting an intruder when in state $s$. We determined the parameters of the error distributions by sampling \num{10000} images at states uniformly distributed throughout the state space and feeding them through the baseline (no risk awareness) perception system to determine whether the intruder is detected. 

\begin{figure}[htb]
    \centering
    \input{daa_em}
    \caption{Perception error model for the vision-based DAA application based on the baseline model. \label{fig:daa_em}}
\end{figure}
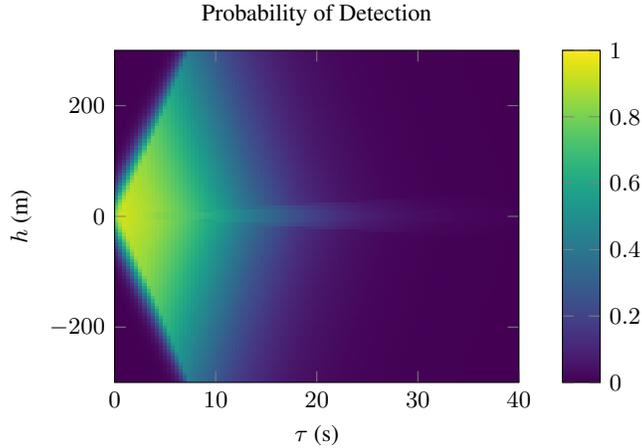

Using the results as training data, we train a small neural network with a single hidden layer of \num{10} hidden and a sigmoid output to represent $\pi_e(\epsilon \mid s)$. We only input part of the state to the neural network. In particular, since $\dot h$, and $a_\text{prev}$ have no influence on the image observation and we only consider cases where $p=1$, we omit those variables at inputs to the error model. Furthermore, noting that the error model should be roughly symmetric about $h=0$, we input $|h|$ rather $h$. In summary, the network takes as input values for $|h|$ and $\tau$ and outputs a probability of detecting the intruder. \Cref{fig:daa_em} shows the resulting model outputs over the state space. The cone shape is due to some states lying outside the ownship field of view.

\textbf{State discretization} \ \ For the distributional dynamic programming, we discretize $h \in [-300, 300]$ and $\dot h \in [-10, 10]$ using \num{41} and \num{21} points respectively with a higher density of points around zero. We discretize $\tau \in [0, 41]$ with a uniform step size of \num{1}. The remaining variables are already discrete.

\textbf{Cost function} \ \ We set the cost function to reflect the separation of the aircraft when $\tau = 0$. Since we want high value of cost to correspond to low separation, we set the cost function as
\begin{equation}
    c(s, \epsilon) = \begin{cases}
    150 - |h|, & \text{if } \tau = 0 \\
    0, & \text{if } \tau > 0
    \end{cases}
\end{equation}
We discretize the cost uniformly between \num{0} and \num{150} using \num{50} points for the distributional dynamic programming.

With these modeling assumptions, we use distributional dynamic programming to solve for $Z^\pi(s, \epsilon)$. Solving for the cost distributions takes under \num{10} seconds on a single Intel Core i7 processor operating at \SI{4.20}{\giga\hertz}. As noted previously, $\dot h$, and $a_\text{prev}$ have no effect on the observed image at a given state, so we marginalize over these variables before calculating the risk function. We compute weights for each discrete value of $\dot h$ and $a_\text{prev}$ from simulations of the controller with perfect perception. We then take the weighted sum of the $Z^\pi(s, a)$ using these weights to obtain a cost distribution function that is only a function of $h$, $\tau$, and $\epsilon$. Using this result, we can compute the risk function.

\subsection{DAA Risk-Driven Data Generation}
To generate data according the samples from the weighting function shown in \cref{fig:daa_rw}, we must map a particular $h$ and $\tau$ into a corresponding image observation. We first sample an ownship position and orientation uniformly in the airspace around the Palo Alto airport (PAO). Next, we select the intruder altitude based on the current altitude of the ownship and the current value for $h$. The remaining position variables are selected based on $\tau$; however, there is not a \num{1}-to-\num{1} mapping due to varying aircraft speeds and relative headings. Therefore, we sample values for the ownship speed, intruder speed, and relative heading and use these values to calculate $\tau$. Specifically, we sample speeds uniformly between \SI{45}{\meter\per\second} and \SI{55}{\meter\per\second} and relative headings between \SI{120}{\degree} and \SI{240}{\degree}. 

\subsection{Encounter Model}\label{app:enc_model}
\textbf{Sampling encounters} For simplicity, the encounters in this work are modeled as straight-line trajectories in which the ownship and intruder follow constant horizontal speeds. We sample an encounter by first sampling a set of encounter features according to uniform distributions over the ranges shown in \cref{tab:enc_model} and then using these features to generate trajectories for the ownship and intruder aircraft.
\begin{table}[htb]
    \centering
    \caption{Encounter model parameters. \label{tab:enc_model}}
    \begin{tabular}{@{}llll@{}}
          \toprule
          \textbf{Parameter} & \textbf{Min} & \textbf{Max} & \textbf{Unit} \\
          \midrule
          Ownship Horizontal Speed & \num{45} & \num{55} & \SI{}{\meter\per\second} \\
          Intruder Horizontal Speed & \num{45} & \num{55} & \SI{}{\meter\per\second} \\
          Horizontal Miss Distance & \num{0} & \num{100} & meters \\
          Vertical Miss Distance & \num{-50} & \num{50} & meters \\
          Relative Heading & \num{120} & \num{240} & degrees \\
          \arrayrulecolor{black}\bottomrule
    \end{tabular}
\end{table}
The horizontal and vertical miss distance parameters indicate the horizontal range and relative altitude of the ownship and intruder aircraft at closest point of approach. We select these distances such that all encounters will results in an NMAC if no collision avoidance action is taken. We simulate encounters that are \SI{50}{\second} long with the closest point of approach occurring \SI{40}{\second} into the encounter. The range of relative headings is set such that the encounters are close to head-on, and the intruder should almost always be within the ownship field of view (although at the beginning of the encounter it may be too small to be detected by the camera). The features in \cref{tab:enc_model} fully determine the relative trajectories of the ownship and intruder. Once the relative trajectories are generated, we rotate and shift both trajectories to put them in random positions around the Palo Alto airport.

\textbf{Simulating encounters} When we simulate the encounters, we add the perception system and ownship controller into the loop. At each time step, we read the horizontal components of the next state from the encounter. For the vertical components, we assume noisy vertical rate centered around zero until the ownship receives an alert, at which point it accelerates to the vertical rate commanded by the collision avoidance advisory.

\subsection{Additional DAA Results}
In addition to simulating each vision-based DAA perception system on the encounter set, we also evaluated the precision and recall of each system on a validation set of \num{1000} images with uniformly sampled intruder positions within the ownship field of view. \Cref{tab:daa_res} shows the results.
\begin{table}[htb]
    \centering
    \caption{Additional DAA results. \label{tab:daa_res}}
    \begin{tabular}{@{}lll@{}}
          \toprule
          & Precision & Recall \\
          \midrule
          Baseline & $0.78 \pm 0.04$ & $0.38 \pm 0.03$ \\
          \arrayrulecolor{black!30}\midrule
          Risk Loss & $0.82 \pm 0.03$ & $0.37 \pm 0.02$ \\
          Risk Data & $0.77 \pm 0.01$ & $0.39 \pm 0.01$ \\
          Both & $0.79 \pm 0.01$ & $0.39 \pm 0.01$ \\
          \arrayrulecolor{black}\bottomrule
    \end{tabular}
\end{table}
We note that one could improve safety of the overall system by simply biasing the perception system towards outputting a detection more often; however, this result would likely increase false positives and result in worse operational efficiency. To ensure that this effect is not causing the safety benefits we show in the risk-driven systems in \cref{fig:daa_results}, we can analyze the effect of risk-driven training on the precision of the overall system. From these results, we determine that the risk-driven design techniques did not harm overall precision and focused their efforts only on risky states.

\end{document}

%% file: system.tex

\tikzset{
    >={Latex[width=2mm,length=2mm]},
        base/.style = {rectangle, rounded corners, draw=black,
        minimum width=1cm, minimum height=1cm,
        text centered},
    block/.style = {base, minimum width=1.2cm, minimum height=0.75cm},
    sutstyle/.style = {block, fill=gray!50},
    envstyle/.style = {block, fill=white}, 
    advstyle/.style = {block, fill=red!15},
}

\begin{tikzpicture}
    [
        node distance=5.5cm,
        every node/.style={font=\small},
        align=center
    ]

    \node (system) [sutstyle] {Controller\\ $u = g(\hat{s})$}; 
    \node (agent) [above of=system, yshift=-4.6cm, xshift=-0.cm] {System};
    \node (perception) [sutstyle, below of=system, yshift=3.5cm] {Perception \\System \\ $\hat{s} = f(o)$}; 
    \node (environment) [envstyle, right of=system, xshift=-1.7cm] {Environment\\$T$}; 

    \draw[->] (system) -- (environment) node [pos=0.55,above] {{\small control $u$}};
    \draw[->] (environment.south)  -- ++(0,-1.6) -- (perception.east) node[pos=0.45,above] {{\small observation $o$}};
    
    \draw[->] (perception) -- (system) node [pos=0.45,left] {{\small $\hat{s}$}};
    \draw[thick,dotted] ($(system.north west)+(-0.3,0.8)$)  rectangle ($(perception.south east)+(0.3,-0.3)$);
\end{tikzpicture}

%% file: adversarial.tex

\tikzset{
    >={Latex[width=2mm,length=2mm]},
        base/.style = {rectangle, rounded corners, draw=black,
        minimum width=1cm, minimum height=1cm,
        text centered},
    block/.style = {base, minimum width=1.2cm, minimum height=0.75cm},
    sutstyle/.style = {block, fill=gray!50},
    envstyle/.style = {block, fill=white}, 
    advstyle/.style = {block, fill=blue!15},
}

\begin{tikzpicture}
    [
        node distance=5.5cm,
        every node/.style={font=\small},
        align=center
    ]

    \node (system) [sutstyle] {Controller\\ $u = g(\hat{s})$}; 
    \node (agent) [above of=system, yshift=-4.6cm, xshift=0.0cm] {$T_\pi$};
    \node (adversary) [advstyle, below of=system, yshift=3.5cm] {Error\\Model\\$\epsilon \sim \pi_e(\cdot \mid s)$}; 
    \node (environment) [envstyle, right of=system, xshift=-1.7cm] {Environment\\$T$}; 

    \draw[->] (system) -- (environment) node [pos=0.45,above] {{\small control $u$}};
    \draw[->] (environment.south)  -- ++(0,-1.6) -- (adversary.east) node[pos=0.45,above] {{\small state $s$}};
    
    \draw[->] (adversary) -- (system) node [pos=0.4,right] {{\small $\hat{s} = s + \epsilon$}};
    \draw[thick,dotted] ($(system.north west)+(-0.3,0.8)$)  rectangle ($(environment.south east)+(0.3,-0.3)$);
    \draw[thick,dotted, white] ($(system.north west)+(-0.3,0.8)$)  rectangle ($(adversary.south east)+(0.3,-0.3)$);
\end{tikzpicture}

%% file: pendulum_overview.tex
\begin{tikzpicture}

\draw (-0.25, -0.25) rectangle (0.25, 0.25);
\draw (0.25, 0.05) -- (0.4, 0.2) -- (0.4, -0.2) -- (0.25, -0.05);
\draw[dotted] (0.4, 0.2) -- (1.4, 1.2);
\draw[dotted] (0.4, -0.2) -- (1.4, -1.2);
\fill[fill = gray!25] (0.4, 0.2) -- (1.4, 1.2) -- (1.4, -1.2) -- (0.4, -0.2) -- cycle;

\fill (1.1, -1.1) circle (0.05);
\draw[thick] (1.1, -1.1) -- (1.1, 1.1);

\draw[->] (1.6, 0.0) -- (2.2, 0.0);

\node at (2.9, 0.1) {\includegraphics[width=1.0cm]{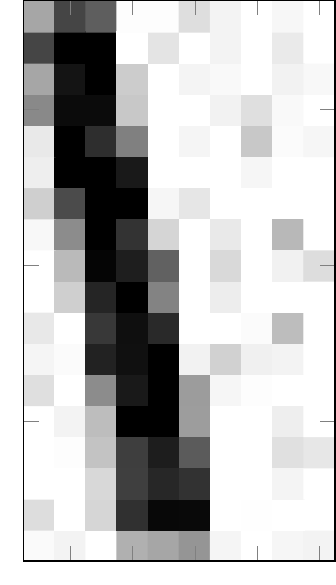}};
\node at (3.1, -0.1) {\includegraphics[width=1.0cm]{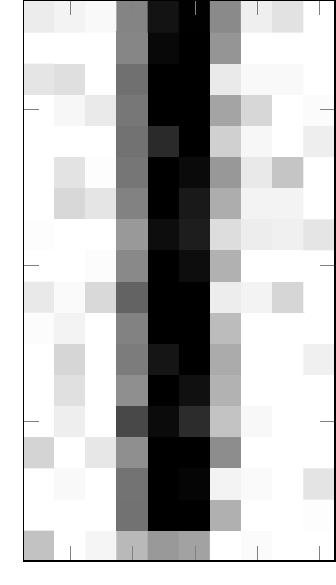}};

\draw[dotted] (-0.4, -1.45) rectangle (3.7, 1.45);
\node at (1.65, 1.65) {\scriptsize Environment};

\draw[->] (3.8, 0.0) -- (4.4, 0.0) node[above, pos=0.5] {\scriptsize $o$};

\node (perception) at (5.5, 0.0) {\scalebox{-1}[1]{\includegraphics[width=2cm]{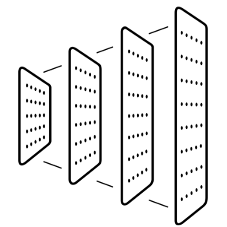}}};
\node[above=of perception, yshift=-1.2cm] {\scriptsize Perception System ($f$)};

\draw[->] (6.7, 0.0) -- (7.3, 0.0) node[above, pos=0.5] {\scriptsize $\hat s$};

\draw (7.5, -0.5) rectangle (9.2, 0.5) node[pos=0.5] {\scriptsize Controller ($\pi$)};

\draw[->] (9.4, 0.0) -- (10.0, 0.0) node[above, pos=0.5] {\scriptsize $a$};

\fill (10.7, -1.1) circle (0.05);
\draw[thick] (10.7, -1.1) -- (10.5, 0.8);
\draw[dashed, opacity=0.5] (10.7, -1.1) -- (10.7, 0.9);
\node at (10.75, -1.0) {\color{red} \LARGE $\curvearrowright$};

\end{tikzpicture}

%% file: pend_cvar_slices_alpha.tex
\begin{tikzpicture}[]

\fill (-2.1, 0.0) circle (0.05);
\draw[very thick] (-2.1, 0) -- (-2.6, 2.45);
\draw[very thick, gray, opacity=0.5] (-2.1, 0) -- (-1.85, 2.5);
\draw[dashed, opacity=0.5] (-2.1, 0.0) -- (-2.1, 2.5);
\node at (-1.85, 2.7) {$\hat s$};
\node at (-2.6, 2.6) {$s$};

\begin{groupplot}[group style={horizontal sep = 0.6cm, vertical sep = 1.5cm, group size=3 by 1}]

\nextgroupplot [
  height = {4.2cm},
  ylabel = {$\epsilon_\omega$},
  title = {$\rho^{\pi_e}_0([0.2, 0.0], \epsilon)$},
  xlabel = {$\epsilon_\theta$},
  width = {4.2cm},
  enlargelimits = false,
  axis on top
]

\addplot[
  point meta min = 0.159,
  point meta max = 0.66
] graphics[
  xmin = -0.4,
  xmax = 0.4,
  ymin = -1,
  ymax = 1
] {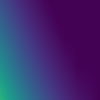};

\nextgroupplot [
  height = {4.2cm},
  title = {$\rho^{\pi_e}_{0.2}([0.2, 0.0], \epsilon)$},
  xlabel = {$\epsilon_\theta$},
  yticklabels = {,},
  width = {4.2cm},
  enlargelimits = false,
  axis on top,
]

\addplot[
  point meta min = 0.159,
  point meta max = 0.66
] graphics[
  xmin = -0.4,
  xmax = 0.4,
  ymin = -1,
  ymax = 1
] {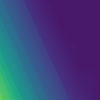};

\nextgroupplot [
  height = {4.2cm},
  title = {$\rho^{\pi_e}_{0.5}([0.2, 0.0], \epsilon)$},
  xlabel = {$\epsilon_\theta$},
  yticklabels = {,},
  width = {4.2cm},
  enlargelimits = false,
  axis on top,
  colormap={mycolormap}{ rgb(0cm)=(0.267004,0.004874,0.329415) rgb(1cm)=(0.26851,0.009605,0.335427) rgb(2cm)=(0.269944,0.014625,0.341379) rgb(3cm)=(0.271305,0.019942,0.347269) rgb(4cm)=(0.272594,0.025563,0.353093) rgb(5cm)=(0.273809,0.031497,0.358853) rgb(6cm)=(0.274952,0.037752,0.364543) rgb(7cm)=(0.276022,0.044167,0.370164) rgb(8cm)=(0.277018,0.050344,0.375715) rgb(9cm)=(0.277941,0.056324,0.381191) rgb(10cm)=(0.278791,0.062145,0.386592) rgb(11cm)=(0.279566,0.067836,0.391917) rgb(12cm)=(0.280267,0.073417,0.397163) rgb(13cm)=(0.280894,0.078907,0.402329) rgb(14cm)=(0.281446,0.08432,0.407414) rgb(15cm)=(0.281924,0.089666,0.412415) rgb(16cm)=(0.282327,0.094955,0.417331) rgb(17cm)=(0.282656,0.100196,0.42216) rgb(18cm)=(0.28291,0.105393,0.426902) rgb(19cm)=(0.283091,0.110553,0.431554) rgb(20cm)=(0.283197,0.11568,0.436115) rgb(21cm)=(0.283229,0.120777,0.440584) rgb(22cm)=(0.283187,0.125848,0.44496) rgb(23cm)=(0.283072,0.130895,0.449241) rgb(24cm)=(0.282884,0.13592,0.453427) rgb(25cm)=(0.282623,0.140926,0.457517) rgb(26cm)=(0.28229,0.145912,0.46151) rgb(27cm)=(0.281887,0.150881,0.465405) rgb(28cm)=(0.281412,0.155834,0.469201) rgb(29cm)=(0.280868,0.160771,0.472899) rgb(30cm)=(0.280255,0.165693,0.476498) rgb(31cm)=(0.279574,0.170599,0.479997) rgb(32cm)=(0.278826,0.17549,0.483397) rgb(33cm)=(0.278012,0.180367,0.486697) rgb(34cm)=(0.277134,0.185228,0.489898) rgb(35cm)=(0.276194,0.190074,0.493001) rgb(36cm)=(0.275191,0.194905,0.496005) rgb(37cm)=(0.274128,0.199721,0.498911) rgb(38cm)=(0.273006,0.20452,0.501721) rgb(39cm)=(0.271828,0.209303,0.504434) rgb(40cm)=(0.270595,0.214069,0.507052) rgb(41cm)=(0.269308,0.218818,0.509577) rgb(42cm)=(0.267968,0.223549,0.512008) rgb(43cm)=(0.26658,0.228262,0.514349) rgb(44cm)=(0.265145,0.232956,0.516599) rgb(45cm)=(0.263663,0.237631,0.518762) rgb(46cm)=(0.262138,0.242286,0.520837) rgb(47cm)=(0.260571,0.246922,0.522828) rgb(48cm)=(0.258965,0.251537,0.524736) rgb(49cm)=(0.257322,0.25613,0.526563) rgb(50cm)=(0.255645,0.260703,0.528312) rgb(51cm)=(0.253935,0.265254,0.529983) rgb(52cm)=(0.252194,0.269783,0.531579) rgb(53cm)=(0.250425,0.27429,0.533103) rgb(54cm)=(0.248629,0.278775,0.534556) rgb(55cm)=(0.246811,0.283237,0.535941) rgb(56cm)=(0.244972,0.287675,0.53726) rgb(57cm)=(0.243113,0.292092,0.538516) rgb(58cm)=(0.241237,0.296485,0.539709) rgb(59cm)=(0.239346,0.300855,0.540844) rgb(60cm)=(0.237441,0.305202,0.541921) rgb(61cm)=(0.235526,0.309527,0.542944) rgb(62cm)=(0.233603,0.313828,0.543914) rgb(63cm)=(0.231674,0.318106,0.544834) rgb(64cm)=(0.229739,0.322361,0.545706) rgb(65cm)=(0.227802,0.326594,0.546532) rgb(66cm)=(0.225863,0.330805,0.547314) rgb(67cm)=(0.223925,0.334994,0.548053) rgb(68cm)=(0.221989,0.339161,0.548752) rgb(69cm)=(0.220057,0.343307,0.549413) rgb(70cm)=(0.21813,0.347432,0.550038) rgb(71cm)=(0.21621,0.351535,0.550627) rgb(72cm)=(0.214298,0.355619,0.551184) rgb(73cm)=(0.212395,0.359683,0.55171) rgb(74cm)=(0.210503,0.363727,0.552206) rgb(75cm)=(0.208623,0.367752,0.552675) rgb(76cm)=(0.206756,0.371758,0.553117) rgb(77cm)=(0.204903,0.375746,0.553533) rgb(78cm)=(0.203063,0.379716,0.553925) rgb(79cm)=(0.201239,0.38367,0.554294) rgb(80cm)=(0.19943,0.387607,0.554642) rgb(81cm)=(0.197636,0.391528,0.554969) rgb(82cm)=(0.19586,0.395433,0.555276) rgb(83cm)=(0.1941,0.399323,0.555565) rgb(84cm)=(0.192357,0.403199,0.555836) rgb(85cm)=(0.190631,0.407061,0.556089) rgb(86cm)=(0.188923,0.41091,0.556326) rgb(87cm)=(0.187231,0.414746,0.556547) rgb(88cm)=(0.185556,0.41857,0.556753) rgb(89cm)=(0.183898,0.422383,0.556944) rgb(90cm)=(0.182256,0.426184,0.55712) rgb(91cm)=(0.180629,0.429975,0.557282) rgb(92cm)=(0.179019,0.433756,0.55743) rgb(93cm)=(0.177423,0.437527,0.557565) rgb(94cm)=(0.175841,0.44129,0.557685) rgb(95cm)=(0.174274,0.445044,0.557792) rgb(96cm)=(0.172719,0.448791,0.557885) rgb(97cm)=(0.171176,0.45253,0.557965) rgb(98cm)=(0.169646,0.456262,0.55803) rgb(99cm)=(0.168126,0.459988,0.558082) rgb(100cm)=(0.166617,0.463708,0.558119) rgb(101cm)=(0.165117,0.467423,0.558141) rgb(102cm)=(0.163625,0.471133,0.558148) rgb(103cm)=(0.162142,0.474838,0.55814) rgb(104cm)=(0.160665,0.47854,0.558115) rgb(105cm)=(0.159194,0.482237,0.558073) rgb(106cm)=(0.157729,0.485932,0.558013) rgb(107cm)=(0.15627,0.489624,0.557936) rgb(108cm)=(0.154815,0.493313,0.55784) rgb(109cm)=(0.153364,0.497,0.557724) rgb(110cm)=(0.151918,0.500685,0.557587) rgb(111cm)=(0.150476,0.504369,0.55743) rgb(112cm)=(0.149039,0.508051,0.55725) rgb(113cm)=(0.147607,0.511733,0.557049) rgb(114cm)=(0.14618,0.515413,0.556823) rgb(115cm)=(0.144759,0.519093,0.556572) rgb(116cm)=(0.143343,0.522773,0.556295) rgb(117cm)=(0.141935,0.526453,0.555991) rgb(118cm)=(0.140536,0.530132,0.555659) rgb(119cm)=(0.139147,0.533812,0.555298) rgb(120cm)=(0.13777,0.537492,0.554906) rgb(121cm)=(0.136408,0.541173,0.554483) rgb(122cm)=(0.135066,0.544853,0.554029) rgb(123cm)=(0.133743,0.548535,0.553541) rgb(124cm)=(0.132444,0.552216,0.553018) rgb(125cm)=(0.131172,0.555899,0.552459) rgb(126cm)=(0.129933,0.559582,0.551864) rgb(127cm)=(0.128729,0.563265,0.551229) rgb(128cm)=(0.127568,0.566949,0.550556) rgb(129cm)=(0.126453,0.570633,0.549841) rgb(130cm)=(0.125394,0.574318,0.549086) rgb(131cm)=(0.124395,0.578002,0.548287) rgb(132cm)=(0.123463,0.581687,0.547445) rgb(133cm)=(0.122606,0.585371,0.546557) rgb(134cm)=(0.121831,0.589055,0.545623) rgb(135cm)=(0.121148,0.592739,0.544641) rgb(136cm)=(0.120565,0.596422,0.543611) rgb(137cm)=(0.120092,0.600104,0.54253) rgb(138cm)=(0.119738,0.603785,0.5414) rgb(139cm)=(0.119512,0.607464,0.540218) rgb(140cm)=(0.119423,0.611141,0.538982) rgb(141cm)=(0.119483,0.614817,0.537692) rgb(142cm)=(0.119699,0.61849,0.536347) rgb(143cm)=(0.120081,0.622161,0.534946) rgb(144cm)=(0.120638,0.625828,0.533488) rgb(145cm)=(0.12138,0.629492,0.531973) rgb(146cm)=(0.122312,0.633153,0.530398) rgb(147cm)=(0.123444,0.636809,0.528763) rgb(148cm)=(0.12478,0.640461,0.527068) rgb(149cm)=(0.126326,0.644107,0.525311) rgb(150cm)=(0.128087,0.647749,0.523491) rgb(151cm)=(0.130067,0.651384,0.521608) rgb(152cm)=(0.132268,0.655014,0.519661) rgb(153cm)=(0.134692,0.658636,0.517649) rgb(154cm)=(0.137339,0.662252,0.515571) rgb(155cm)=(0.14021,0.665859,0.513427) rgb(156cm)=(0.143303,0.669459,0.511215) rgb(157cm)=(0.146616,0.67305,0.508936) rgb(158cm)=(0.150148,0.676631,0.506589) rgb(159cm)=(0.153894,0.680203,0.504172) rgb(160cm)=(0.157851,0.683765,0.501686) rgb(161cm)=(0.162016,0.687316,0.499129) rgb(162cm)=(0.166383,0.690856,0.496502) rgb(163cm)=(0.170948,0.694384,0.493803) rgb(164cm)=(0.175707,0.6979,0.491033) rgb(165cm)=(0.180653,0.701402,0.488189) rgb(166cm)=(0.185783,0.704891,0.485273) rgb(167cm)=(0.19109,0.708366,0.482284) rgb(168cm)=(0.196571,0.711827,0.479221) rgb(169cm)=(0.202219,0.715272,0.476084) rgb(170cm)=(0.20803,0.718701,0.472873) rgb(171cm)=(0.214,0.722114,0.469588) rgb(172cm)=(0.220124,0.725509,0.466226) rgb(173cm)=(0.226397,0.728888,0.462789) rgb(174cm)=(0.232815,0.732247,0.459277) rgb(175cm)=(0.239374,0.735588,0.455688) rgb(176cm)=(0.24607,0.73891,0.452024) rgb(177cm)=(0.252899,0.742211,0.448284) rgb(178cm)=(0.259857,0.745492,0.444467) rgb(179cm)=(0.266941,0.748751,0.440573) rgb(180cm)=(0.274149,0.751988,0.436601) rgb(181cm)=(0.281477,0.755203,0.432552) rgb(182cm)=(0.288921,0.758394,0.428426) rgb(183cm)=(0.296479,0.761561,0.424223) rgb(184cm)=(0.304148,0.764704,0.419943) rgb(185cm)=(0.311925,0.767822,0.415586) rgb(186cm)=(0.319809,0.770914,0.411152) rgb(187cm)=(0.327796,0.77398,0.40664) rgb(188cm)=(0.335885,0.777018,0.402049) rgb(189cm)=(0.344074,0.780029,0.397381) rgb(190cm)=(0.35236,0.783011,0.392636) rgb(191cm)=(0.360741,0.785964,0.387814) rgb(192cm)=(0.369214,0.788888,0.382914) rgb(193cm)=(0.377779,0.791781,0.377939) rgb(194cm)=(0.386433,0.794644,0.372886) rgb(195cm)=(0.395174,0.797475,0.367757) rgb(196cm)=(0.404001,0.800275,0.362552) rgb(197cm)=(0.412913,0.803041,0.357269) rgb(198cm)=(0.421908,0.805774,0.35191) rgb(199cm)=(0.430983,0.808473,0.346476) rgb(200cm)=(0.440137,0.811138,0.340967) rgb(201cm)=(0.449368,0.813768,0.335384) rgb(202cm)=(0.458674,0.816363,0.329727) rgb(203cm)=(0.468053,0.818921,0.323998) rgb(204cm)=(0.477504,0.821444,0.318195) rgb(205cm)=(0.487026,0.823929,0.312321) rgb(206cm)=(0.496615,0.826376,0.306377) rgb(207cm)=(0.506271,0.828786,0.300362) rgb(208cm)=(0.515992,0.831158,0.294279) rgb(209cm)=(0.525776,0.833491,0.288127) rgb(210cm)=(0.535621,0.835785,0.281908) rgb(211cm)=(0.545524,0.838039,0.275626) rgb(212cm)=(0.555484,0.840254,0.269281) rgb(213cm)=(0.565498,0.84243,0.262877) rgb(214cm)=(0.575563,0.844566,0.256415) rgb(215cm)=(0.585678,0.846661,0.249897) rgb(216cm)=(0.595839,0.848717,0.243329) rgb(217cm)=(0.606045,0.850733,0.236712) rgb(218cm)=(0.616293,0.852709,0.230052) rgb(219cm)=(0.626579,0.854645,0.223353) rgb(220cm)=(0.636902,0.856542,0.21662) rgb(221cm)=(0.647257,0.8584,0.209861) rgb(222cm)=(0.657642,0.860219,0.203082) rgb(223cm)=(0.668054,0.861999,0.196293) rgb(224cm)=(0.678489,0.863742,0.189503) rgb(225cm)=(0.688944,0.865448,0.182725) rgb(226cm)=(0.699415,0.867117,0.175971) rgb(227cm)=(0.709898,0.868751,0.169257) rgb(228cm)=(0.720391,0.87035,0.162603) rgb(229cm)=(0.730889,0.871916,0.156029) rgb(230cm)=(0.741388,0.873449,0.149561) rgb(231cm)=(0.751884,0.874951,0.143228) rgb(232cm)=(0.762373,0.876424,0.137064) rgb(233cm)=(0.772852,0.877868,0.131109) rgb(234cm)=(0.783315,0.879285,0.125405) rgb(235cm)=(0.79376,0.880678,0.120005) rgb(236cm)=(0.804182,0.882046,0.114965) rgb(237cm)=(0.814576,0.883393,0.110347) rgb(238cm)=(0.82494,0.88472,0.106217) rgb(239cm)=(0.83527,0.886029,0.102646) rgb(240cm)=(0.845561,0.887322,0.099702) rgb(241cm)=(0.85581,0.888601,0.097452) rgb(242cm)=(0.866013,0.889868,0.095953) rgb(243cm)=(0.876168,0.891125,0.09525) rgb(244cm)=(0.886271,0.892374,0.095374) rgb(245cm)=(0.89632,0.893616,0.096335) rgb(246cm)=(0.906311,0.894855,0.098125) rgb(247cm)=(0.916242,0.896091,0.100717) rgb(248cm)=(0.926106,0.89733,0.104071) rgb(249cm)=(0.935904,0.89857,0.108131) rgb(250cm)=(0.945636,0.899815,0.112838) rgb(251cm)=(0.9553,0.901065,0.118128) rgb(252cm)=(0.964894,0.902323,0.123941) rgb(253cm)=(0.974417,0.90359,0.130215) rgb(254cm)=(0.983868,0.904867,0.136897) rgb(255cm)=(0.993248,0.906157,0.143936) },
  colorbar
]

\addplot[
  point meta min = 0.159,
  point meta max = 0.66
] graphics[
  xmin = -0.4,
  xmax = 0.4,
  ymin = -1,
  ymax = 1
] {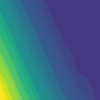};

\end{groupplot}

\draw[->] (0.25, 0.15) -- (-1.9, 1.0);

\end{tikzpicture}

%% file: pend_rw.tex
\begin{tikzpicture}[]
\begin{groupplot}[group style={horizontal sep = 0.5cm, vertical sep = 1.5cm, group size=3 by 1}]

\nextgroupplot [
  height = {4.2cm},
  ylabel = {$\omega$ (rad/s)},
  xlabel = {$\theta$ (rad)},
  title = {$\max_{\epsilon \in \mathcal E} \rho^{\pi_e}_0(s, \epsilon)$},
  width = {4.2cm},
  enlargelimits = false,
  axis on top
]

\addplot[
  point meta min = 0.0,
  point meta max = 0.8
] graphics[
  xmin = -0.7853981633974483,
  xmax = 0.7853981633974483,
  ymin = -2,
  ymax = 2
] {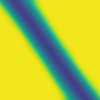};

\nextgroupplot [
  height = {4.2cm},
  xlabel = {$\theta$ (rad)},
  yticklabels = {,},
  title = {$\rho^{\pi_e}_0(s, 0)$},
  width = {4.2cm},
  enlargelimits = false,
  axis on top
]

\addplot[
  point meta min = 0.0,
  point meta max = 0.8
] graphics[
  xmin = -0.7853981633974483,
  xmax = 0.7853981633974483,
  ymin = -2,
  ymax = 2
] {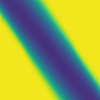};

\nextgroupplot [
  height = {4.2cm},
  xlabel = {$\theta$ (rad)},
  title = {$w_0(s)$},
  yticklabels = {,},
  width = {4.2cm},
  enlargelimits = false,
  axis on top,
  colormap={mycolormap}{ rgb(0cm)=(0.267004,0.004874,0.329415) rgb(1cm)=(0.26851,0.009605,0.335427) rgb(2cm)=(0.269944,0.014625,0.341379) rgb(3cm)=(0.271305,0.019942,0.347269) rgb(4cm)=(0.272594,0.025563,0.353093) rgb(5cm)=(0.273809,0.031497,0.358853) rgb(6cm)=(0.274952,0.037752,0.364543) rgb(7cm)=(0.276022,0.044167,0.370164) rgb(8cm)=(0.277018,0.050344,0.375715) rgb(9cm)=(0.277941,0.056324,0.381191) rgb(10cm)=(0.278791,0.062145,0.386592) rgb(11cm)=(0.279566,0.067836,0.391917) rgb(12cm)=(0.280267,0.073417,0.397163) rgb(13cm)=(0.280894,0.078907,0.402329) rgb(14cm)=(0.281446,0.08432,0.407414) rgb(15cm)=(0.281924,0.089666,0.412415) rgb(16cm)=(0.282327,0.094955,0.417331) rgb(17cm)=(0.282656,0.100196,0.42216) rgb(18cm)=(0.28291,0.105393,0.426902) rgb(19cm)=(0.283091,0.110553,0.431554) rgb(20cm)=(0.283197,0.11568,0.436115) rgb(21cm)=(0.283229,0.120777,0.440584) rgb(22cm)=(0.283187,0.125848,0.44496) rgb(23cm)=(0.283072,0.130895,0.449241) rgb(24cm)=(0.282884,0.13592,0.453427) rgb(25cm)=(0.282623,0.140926,0.457517) rgb(26cm)=(0.28229,0.145912,0.46151) rgb(27cm)=(0.281887,0.150881,0.465405) rgb(28cm)=(0.281412,0.155834,0.469201) rgb(29cm)=(0.280868,0.160771,0.472899) rgb(30cm)=(0.280255,0.165693,0.476498) rgb(31cm)=(0.279574,0.170599,0.479997) rgb(32cm)=(0.278826,0.17549,0.483397) rgb(33cm)=(0.278012,0.180367,0.486697) rgb(34cm)=(0.277134,0.185228,0.489898) rgb(35cm)=(0.276194,0.190074,0.493001) rgb(36cm)=(0.275191,0.194905,0.496005) rgb(37cm)=(0.274128,0.199721,0.498911) rgb(38cm)=(0.273006,0.20452,0.501721) rgb(39cm)=(0.271828,0.209303,0.504434) rgb(40cm)=(0.270595,0.214069,0.507052) rgb(41cm)=(0.269308,0.218818,0.509577) rgb(42cm)=(0.267968,0.223549,0.512008) rgb(43cm)=(0.26658,0.228262,0.514349) rgb(44cm)=(0.265145,0.232956,0.516599) rgb(45cm)=(0.263663,0.237631,0.518762) rgb(46cm)=(0.262138,0.242286,0.520837) rgb(47cm)=(0.260571,0.246922,0.522828) rgb(48cm)=(0.258965,0.251537,0.524736) rgb(49cm)=(0.257322,0.25613,0.526563) rgb(50cm)=(0.255645,0.260703,0.528312) rgb(51cm)=(0.253935,0.265254,0.529983) rgb(52cm)=(0.252194,0.269783,0.531579) rgb(53cm)=(0.250425,0.27429,0.533103) rgb(54cm)=(0.248629,0.278775,0.534556) rgb(55cm)=(0.246811,0.283237,0.535941) rgb(56cm)=(0.244972,0.287675,0.53726) rgb(57cm)=(0.243113,0.292092,0.538516) rgb(58cm)=(0.241237,0.296485,0.539709) rgb(59cm)=(0.239346,0.300855,0.540844) rgb(60cm)=(0.237441,0.305202,0.541921) rgb(61cm)=(0.235526,0.309527,0.542944) rgb(62cm)=(0.233603,0.313828,0.543914) rgb(63cm)=(0.231674,0.318106,0.544834) rgb(64cm)=(0.229739,0.322361,0.545706) rgb(65cm)=(0.227802,0.326594,0.546532) rgb(66cm)=(0.225863,0.330805,0.547314) rgb(67cm)=(0.223925,0.334994,0.548053) rgb(68cm)=(0.221989,0.339161,0.548752) rgb(69cm)=(0.220057,0.343307,0.549413) rgb(70cm)=(0.21813,0.347432,0.550038) rgb(71cm)=(0.21621,0.351535,0.550627) rgb(72cm)=(0.214298,0.355619,0.551184) rgb(73cm)=(0.212395,0.359683,0.55171) rgb(74cm)=(0.210503,0.363727,0.552206) rgb(75cm)=(0.208623,0.367752,0.552675) rgb(76cm)=(0.206756,0.371758,0.553117) rgb(77cm)=(0.204903,0.375746,0.553533) rgb(78cm)=(0.203063,0.379716,0.553925) rgb(79cm)=(0.201239,0.38367,0.554294) rgb(80cm)=(0.19943,0.387607,0.554642) rgb(81cm)=(0.197636,0.391528,0.554969) rgb(82cm)=(0.19586,0.395433,0.555276) rgb(83cm)=(0.1941,0.399323,0.555565) rgb(84cm)=(0.192357,0.403199,0.555836) rgb(85cm)=(0.190631,0.407061,0.556089) rgb(86cm)=(0.188923,0.41091,0.556326) rgb(87cm)=(0.187231,0.414746,0.556547) rgb(88cm)=(0.185556,0.41857,0.556753) rgb(89cm)=(0.183898,0.422383,0.556944) rgb(90cm)=(0.182256,0.426184,0.55712) rgb(91cm)=(0.180629,0.429975,0.557282) rgb(92cm)=(0.179019,0.433756,0.55743) rgb(93cm)=(0.177423,0.437527,0.557565) rgb(94cm)=(0.175841,0.44129,0.557685) rgb(95cm)=(0.174274,0.445044,0.557792) rgb(96cm)=(0.172719,0.448791,0.557885) rgb(97cm)=(0.171176,0.45253,0.557965) rgb(98cm)=(0.169646,0.456262,0.55803) rgb(99cm)=(0.168126,0.459988,0.558082) rgb(100cm)=(0.166617,0.463708,0.558119) rgb(101cm)=(0.165117,0.467423,0.558141) rgb(102cm)=(0.163625,0.471133,0.558148) rgb(103cm)=(0.162142,0.474838,0.55814) rgb(104cm)=(0.160665,0.47854,0.558115) rgb(105cm)=(0.159194,0.482237,0.558073) rgb(106cm)=(0.157729,0.485932,0.558013) rgb(107cm)=(0.15627,0.489624,0.557936) rgb(108cm)=(0.154815,0.493313,0.55784) rgb(109cm)=(0.153364,0.497,0.557724) rgb(110cm)=(0.151918,0.500685,0.557587) rgb(111cm)=(0.150476,0.504369,0.55743) rgb(112cm)=(0.149039,0.508051,0.55725) rgb(113cm)=(0.147607,0.511733,0.557049) rgb(114cm)=(0.14618,0.515413,0.556823) rgb(115cm)=(0.144759,0.519093,0.556572) rgb(116cm)=(0.143343,0.522773,0.556295) rgb(117cm)=(0.141935,0.526453,0.555991) rgb(118cm)=(0.140536,0.530132,0.555659) rgb(119cm)=(0.139147,0.533812,0.555298) rgb(120cm)=(0.13777,0.537492,0.554906) rgb(121cm)=(0.136408,0.541173,0.554483) rgb(122cm)=(0.135066,0.544853,0.554029) rgb(123cm)=(0.133743,0.548535,0.553541) rgb(124cm)=(0.132444,0.552216,0.553018) rgb(125cm)=(0.131172,0.555899,0.552459) rgb(126cm)=(0.129933,0.559582,0.551864) rgb(127cm)=(0.128729,0.563265,0.551229) rgb(128cm)=(0.127568,0.566949,0.550556) rgb(129cm)=(0.126453,0.570633,0.549841) rgb(130cm)=(0.125394,0.574318,0.549086) rgb(131cm)=(0.124395,0.578002,0.548287) rgb(132cm)=(0.123463,0.581687,0.547445) rgb(133cm)=(0.122606,0.585371,0.546557) rgb(134cm)=(0.121831,0.589055,0.545623) rgb(135cm)=(0.121148,0.592739,0.544641) rgb(136cm)=(0.120565,0.596422,0.543611) rgb(137cm)=(0.120092,0.600104,0.54253) rgb(138cm)=(0.119738,0.603785,0.5414) rgb(139cm)=(0.119512,0.607464,0.540218) rgb(140cm)=(0.119423,0.611141,0.538982) rgb(141cm)=(0.119483,0.614817,0.537692) rgb(142cm)=(0.119699,0.61849,0.536347) rgb(143cm)=(0.120081,0.622161,0.534946) rgb(144cm)=(0.120638,0.625828,0.533488) rgb(145cm)=(0.12138,0.629492,0.531973) rgb(146cm)=(0.122312,0.633153,0.530398) rgb(147cm)=(0.123444,0.636809,0.528763) rgb(148cm)=(0.12478,0.640461,0.527068) rgb(149cm)=(0.126326,0.644107,0.525311) rgb(150cm)=(0.128087,0.647749,0.523491) rgb(151cm)=(0.130067,0.651384,0.521608) rgb(152cm)=(0.132268,0.655014,0.519661) rgb(153cm)=(0.134692,0.658636,0.517649) rgb(154cm)=(0.137339,0.662252,0.515571) rgb(155cm)=(0.14021,0.665859,0.513427) rgb(156cm)=(0.143303,0.669459,0.511215) rgb(157cm)=(0.146616,0.67305,0.508936) rgb(158cm)=(0.150148,0.676631,0.506589) rgb(159cm)=(0.153894,0.680203,0.504172) rgb(160cm)=(0.157851,0.683765,0.501686) rgb(161cm)=(0.162016,0.687316,0.499129) rgb(162cm)=(0.166383,0.690856,0.496502) rgb(163cm)=(0.170948,0.694384,0.493803) rgb(164cm)=(0.175707,0.6979,0.491033) rgb(165cm)=(0.180653,0.701402,0.488189) rgb(166cm)=(0.185783,0.704891,0.485273) rgb(167cm)=(0.19109,0.708366,0.482284) rgb(168cm)=(0.196571,0.711827,0.479221) rgb(169cm)=(0.202219,0.715272,0.476084) rgb(170cm)=(0.20803,0.718701,0.472873) rgb(171cm)=(0.214,0.722114,0.469588) rgb(172cm)=(0.220124,0.725509,0.466226) rgb(173cm)=(0.226397,0.728888,0.462789) rgb(174cm)=(0.232815,0.732247,0.459277) rgb(175cm)=(0.239374,0.735588,0.455688) rgb(176cm)=(0.24607,0.73891,0.452024) rgb(177cm)=(0.252899,0.742211,0.448284) rgb(178cm)=(0.259857,0.745492,0.444467) rgb(179cm)=(0.266941,0.748751,0.440573) rgb(180cm)=(0.274149,0.751988,0.436601) rgb(181cm)=(0.281477,0.755203,0.432552) rgb(182cm)=(0.288921,0.758394,0.428426) rgb(183cm)=(0.296479,0.761561,0.424223) rgb(184cm)=(0.304148,0.764704,0.419943) rgb(185cm)=(0.311925,0.767822,0.415586) rgb(186cm)=(0.319809,0.770914,0.411152) rgb(187cm)=(0.327796,0.77398,0.40664) rgb(188cm)=(0.335885,0.777018,0.402049) rgb(189cm)=(0.344074,0.780029,0.397381) rgb(190cm)=(0.35236,0.783011,0.392636) rgb(191cm)=(0.360741,0.785964,0.387814) rgb(192cm)=(0.369214,0.788888,0.382914) rgb(193cm)=(0.377779,0.791781,0.377939) rgb(194cm)=(0.386433,0.794644,0.372886) rgb(195cm)=(0.395174,0.797475,0.367757) rgb(196cm)=(0.404001,0.800275,0.362552) rgb(197cm)=(0.412913,0.803041,0.357269) rgb(198cm)=(0.421908,0.805774,0.35191) rgb(199cm)=(0.430983,0.808473,0.346476) rgb(200cm)=(0.440137,0.811138,0.340967) rgb(201cm)=(0.449368,0.813768,0.335384) rgb(202cm)=(0.458674,0.816363,0.329727) rgb(203cm)=(0.468053,0.818921,0.323998) rgb(204cm)=(0.477504,0.821444,0.318195) rgb(205cm)=(0.487026,0.823929,0.312321) rgb(206cm)=(0.496615,0.826376,0.306377) rgb(207cm)=(0.506271,0.828786,0.300362) rgb(208cm)=(0.515992,0.831158,0.294279) rgb(209cm)=(0.525776,0.833491,0.288127) rgb(210cm)=(0.535621,0.835785,0.281908) rgb(211cm)=(0.545524,0.838039,0.275626) rgb(212cm)=(0.555484,0.840254,0.269281) rgb(213cm)=(0.565498,0.84243,0.262877) rgb(214cm)=(0.575563,0.844566,0.256415) rgb(215cm)=(0.585678,0.846661,0.249897) rgb(216cm)=(0.595839,0.848717,0.243329) rgb(217cm)=(0.606045,0.850733,0.236712) rgb(218cm)=(0.616293,0.852709,0.230052) rgb(219cm)=(0.626579,0.854645,0.223353) rgb(220cm)=(0.636902,0.856542,0.21662) rgb(221cm)=(0.647257,0.8584,0.209861) rgb(222cm)=(0.657642,0.860219,0.203082) rgb(223cm)=(0.668054,0.861999,0.196293) rgb(224cm)=(0.678489,0.863742,0.189503) rgb(225cm)=(0.688944,0.865448,0.182725) rgb(226cm)=(0.699415,0.867117,0.175971) rgb(227cm)=(0.709898,0.868751,0.169257) rgb(228cm)=(0.720391,0.87035,0.162603) rgb(229cm)=(0.730889,0.871916,0.156029) rgb(230cm)=(0.741388,0.873449,0.149561) rgb(231cm)=(0.751884,0.874951,0.143228) rgb(232cm)=(0.762373,0.876424,0.137064) rgb(233cm)=(0.772852,0.877868,0.131109) rgb(234cm)=(0.783315,0.879285,0.125405) rgb(235cm)=(0.79376,0.880678,0.120005) rgb(236cm)=(0.804182,0.882046,0.114965) rgb(237cm)=(0.814576,0.883393,0.110347) rgb(238cm)=(0.82494,0.88472,0.106217) rgb(239cm)=(0.83527,0.886029,0.102646) rgb(240cm)=(0.845561,0.887322,0.099702) rgb(241cm)=(0.85581,0.888601,0.097452) rgb(242cm)=(0.866013,0.889868,0.095953) rgb(243cm)=(0.876168,0.891125,0.09525) rgb(244cm)=(0.886271,0.892374,0.095374) rgb(245cm)=(0.89632,0.893616,0.096335) rgb(246cm)=(0.906311,0.894855,0.098125) rgb(247cm)=(0.916242,0.896091,0.100717) rgb(248cm)=(0.926106,0.89733,0.104071) rgb(249cm)=(0.935904,0.89857,0.108131) rgb(250cm)=(0.945636,0.899815,0.112838) rgb(251cm)=(0.9553,0.901065,0.118128) rgb(252cm)=(0.964894,0.902323,0.123941) rgb(253cm)=(0.974417,0.90359,0.130215) rgb(254cm)=(0.983868,0.904867,0.136897) rgb(255cm)=(0.993248,0.906157,0.143936) },
  colorbar
]

\addplot[
  point meta min = 0.0,
  point meta max = 0.8
] graphics[
  xmin = -0.7853981633974483,
  xmax = 0.7853981633974483,
  ymin = -2,
  ymax = 2
] {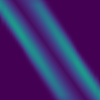};

\end{groupplot}

\end{tikzpicture}

%% file: daa_overview.tex
\begin{tikzpicture}












\node (ownship) [aircraft side,fill=black,draw=black,minimum width=1cm,scale=0.5] at (0,0) {};

\draw[dotted] (0.15, 0.0) -- (2.0, 1.5);
\draw[dotted] (0.15, 0.0) -- (2.0, -1.5);
\fill[gray!25] (0.15, 0.0) -- (2.0, 1.5) -- (2.0, -1.5) -- cycle;

\node (intruder) [aircraft side,fill=black,xscale=-1,draw=black,minimum width=1cm,scale=0.5] at (1.5, 0.5) {};

\node at (0, 0.25) {\scriptsize ownship};
\node at (1.5, 0.75) {\scriptsize intruder};

\draw[->] (2.2, 0.0) -- (2.8, 0.0);

\node at (4.1, 0.0) {\includegraphics[trim={15cm 0 15cm 0},clip,width=2.2cm]{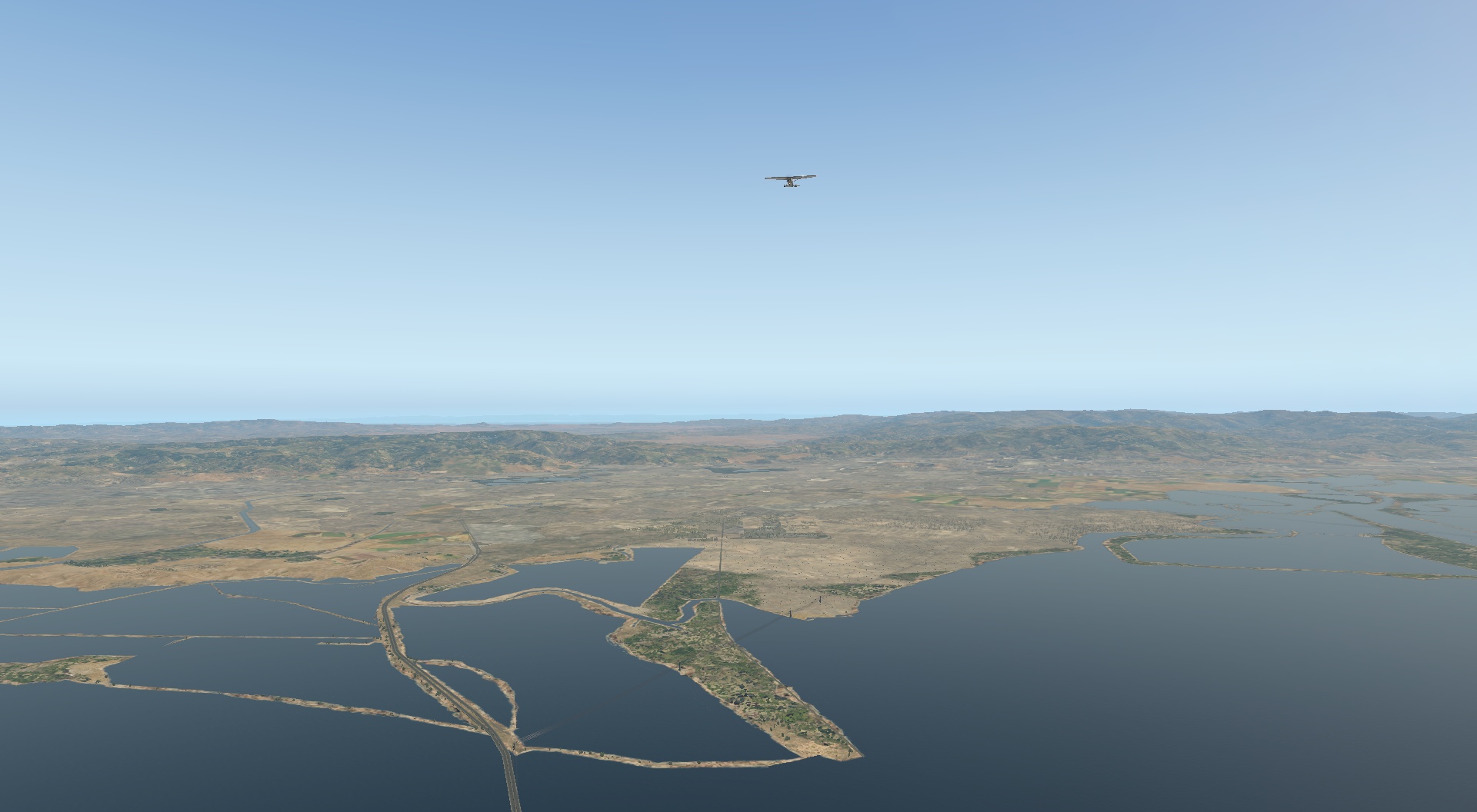}};

\draw[dotted] (-0.5, -1.7) rectangle (5.4, 1.7);
\node at (2.45, 1.9) {\scriptsize Environment};

\draw[->] (5.6, 0.0) -- (6.2, 0.0);
\node at (5.9, 0.15) {\scriptsize $o$};

\node (perception) at (7.5, 0.0) {\includegraphics[trim={15cm 0 15cm 0},clip,width=2.2cm]{images/daa_im_ex_v2.jpg}};
\draw[red] (7.52, 0.55) rectangle (7.72, 0.65);
\node[above=of perception, yshift=-1.2cm] {\scriptsize Perception System ($f$)};

\draw[->] (8.8, 0.0) -- (9.4, 0.0);
\node at (9.1, 0.2) {\scriptsize $\hat s$};

\node (ownship) [aircraft side,fill=black,draw=black,minimum width=1cm,scale=0.5] at (10,0) {};

\draw[dotted] (10.15, 0.0) -- (12.0, 1.5);
\draw[dotted] (10.15, 0.0) -- (12.0, -1.5);
\fill[gray!25] (10.15, 0.0) -- (12.0, 1.5) -- (12.0, -1.5) -- cycle;

\node (intruder) [aircraft side,fill=black,xscale=-1,draw=black,minimum width=1cm,scale=0.5] at (11.5, 0.5) {};

\draw[red] (10.15, 0.0) -- (10.3, 0.0);
\draw[->, red] (10.3, 0.0) to [bend left=45] (10.7, -0.4);

\node at (11.0,1.7) {\scriptsize Controller $g$};

\end{tikzpicture}

%% file: daa_rw.tex
\begin{tikzpicture}[]






\begin{axis}[
  height = {4.5cm},
  ylabel = {$h$ (m)},
  xlabel = {$\tau$ (s)},
  width = {4.5cm},
  enlargelimits = false,
  axis on top,
]

\addplot+[
  point meta min = 0.0,
  point meta max = 11.97861634250583
] graphics[
  xmin = 0,
  xmax = 40,
  ymin = -300,
  ymax = 300
] {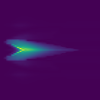};

\end{axis}

\end{tikzpicture}

%% file: daa_results_v2.tex
\begin{tikzpicture}[]
\begin{groupplot}[group style={horizontal sep = 1cm, group size=2 by 1}]

\nextgroupplot [
  ybar = 0pt,
  bar width = 8pt,
  xtick = data,
  symbolic x coords = {a},
  xticklabels={,},
  xmajorticks={false},
  xticklabel style={align=center},
  ylabel = {No. of NMACs},
  ymin = {0},
  ymax = {600},
  width = {3.5cm},
  height = {4cm},
  enlarge x limits=0.25,
]

\addplot+[
  gray, draw=gray, fill=gray!50,
  error bars/.cd,
  x dir=both,
  x explicit,
  y dir=both,
  y explicit
] table[
  x error plus=ex+,
  x error minus=ex-,
  y error plus=ey+,
  y error minus=ey-
] {
  x y ex+ ex- ey+ ey-
  a 504 0.0 0.0 9.6 9.6
};

\addplot+[
  pastelBlue, draw=pastelBlue, fill=pastelBlue!50,
  error bars/.cd,
  x dir=both,
  x explicit,
  y dir=both,
  y explicit
] table[
  x error plus=ex+,
  x error minus=ex-,
  y error plus=ey+,
  y error minus=ey-
] {
  x y ex+ ex- ey+ ey-
  a 443 0.0 0.0 6.2 6.2
};

\addplot+[
  pastelRed, draw=pastelRed, fill=pastelRed!50,
  error bars/.cd,
  x dir=both,
  x explicit,
  y dir=both,
  y explicit
] table[
  x error plus=ex+,
  x error minus=ex-,
  y error plus=ey+,
  y error minus=ey-
] {
  x y ex+ ex- ey+ ey-
  a 337 0.0 0.0 17 17
};

\addplot+[
  pastelPurple, draw=pastelPurple, fill=pastelPurple!50,
  error bars/.cd,
  x dir=both,
  x explicit,
  y dir=both,
  y explicit
] table[
  x error plus=ex+,
  x error minus=ex-,
  y error plus=ey+,
  y error minus=ey-
] {
  x y ex+ ex- ey+ ey-
  a 317 0.0 0.0 8.3 8.3
};

\nextgroupplot [
  xmin = {0.0},
  xmax = {150.0},
  ymax = {1.0},
  xlabel = {Risk},
  ymin = {0.0},
  width = {8cm},
  height = {4cm},
  legend columns=-1,
  legend style = {font=\footnotesize, at={(-0.33,1.35)}, anchor=north west}
]

\addplot+[
  gray, mark = {none}
] coordinates {
  (0.0, 0.0)
  (1.0, 0.0)
  (2.0, 0.0)
  (3.0, 0.0)
  (4.0, 0.0)
  (5.0, 9.333333333333333e-5)
  (6.0, 0.00012)
  (7.0, 0.00016)
  (8.0, 0.0002866666666666667)
  (9.0, 0.00042)
  (10.0, 0.0005)
  (11.0, 0.0005733333333333334)
  (12.0, 0.0005866666666666667)
  (13.0, 0.0006066666666666667)
  (14.0, 0.0006466666666666667)
  (15.0, 0.00076)
  (16.0, 0.0008133333333333333)
  (17.0, 0.0008866666666666667)
  (18.0, 0.0011)
  (19.0, 0.0011733333333333333)
  (20.0, 0.0015666666666666667)
  (21.0, 0.0018533333333333334)
  (22.0, 0.0021666666666666666)
  (23.0, 0.00236)
  (24.0, 0.0025066666666666666)
  (25.0, 0.0029)
  (26.0, 0.0032066666666666667)
  (27.0, 0.0035666666666666668)
  (28.0, 0.0037333333333333333)
  (29.0, 0.004026666666666666)
  (30.0, 0.004573333333333334)
  (31.0, 0.005066666666666666)
  (32.0, 0.0057866666666666665)
  (33.0, 0.006486666666666667)
  (34.0, 0.007153333333333334)
  (35.0, 0.008106666666666667)
  (36.0, 0.00842)
  (37.0, 0.008866666666666667)
  (38.0, 0.009706666666666667)
  (39.0, 0.01066)
  (40.0, 0.01144)
  (41.0, 0.012573333333333334)
  (42.0, 0.01326)
  (43.0, 0.014066666666666667)
  (44.0, 0.014853333333333333)
  (45.0, 0.015666666666666666)
  (46.0, 0.01688)
  (47.0, 0.017833333333333333)
  (48.0, 0.01924)
  (49.0, 0.020133333333333333)
  (50.0, 0.021466666666666665)
  (51.0, 0.022493333333333334)
  (52.0, 0.023773333333333334)
  (53.0, 0.02492)
  (54.0, 0.026513333333333333)
  (55.0, 0.028046666666666668)
  (56.0, 0.030153333333333334)
  (57.0, 0.03194)
  (58.0, 0.03360666666666667)
  (59.0, 0.035693333333333334)
  (60.0, 0.03754)
  (61.0, 0.03897333333333333)
  (62.0, 0.040806666666666665)
  (63.0, 0.04269333333333333)
  (64.0, 0.04504)
  (65.0, 0.04751333333333333)
  (66.0, 0.04972)
  (67.0, 0.05213333333333333)
  (68.0, 0.05481333333333333)
  (69.0, 0.05767333333333333)
  (70.0, 0.0607)
  (71.0, 0.06428)
  (72.0, 0.06852)
  (73.0, 0.07206666666666667)
  (74.0, 0.07581333333333333)
  (75.0, 0.07940666666666667)
  (76.0, 0.08422)
  (77.0, 0.08910666666666667)
  (78.0, 0.09433333333333334)
  (79.0, 0.10049333333333334)
  (80.0, 0.12919333333333333)
  (81.0, 0.18931333333333333)
  (82.0, 0.23946666666666666)
  (83.0, 0.27706)
  (84.0, 0.31315333333333334)
  (85.0, 0.34747333333333336)
  (86.0, 0.37432)
  (87.0, 0.40047333333333335)
  (88.0, 0.42485333333333336)
  (89.0, 0.44896)
  (90.0, 0.4736066666666667)
  (91.0, 0.49634666666666666)
  (92.0, 0.5177466666666667)
  (93.0, 0.5402666666666667)
  (94.0, 0.5656266666666667)
  (95.0, 0.5851733333333333)
  (96.0, 0.6065)
  (97.0, 0.6283133333333333)
  (98.0, 0.6526266666666667)
  (99.0, 0.6734866666666667)
  (100.0, 0.69454)
  (101.0, 0.7171866666666666)
  (102.0, 0.7386133333333333)
  (103.0, 0.7569933333333333)
  (104.0, 0.7769866666666667)
  (105.0, 0.7976266666666667)
  (106.0, 0.8155533333333334)
  (107.0, 0.8348)
  (108.0, 0.8510133333333333)
  (109.0, 0.8678266666666666)
  (110.0, 0.8847466666666667)
  (111.0, 0.90018)
  (112.0, 0.9108133333333334)
  (113.0, 0.9190733333333333)
  (114.0, 0.9257066666666667)
  (115.0, 0.9316133333333333)
  (116.0, 0.93728)
  (117.0, 0.9421)
  (118.0, 0.9472866666666667)
  (119.0, 0.9517533333333333)
  (120.0, 0.9558466666666666)
  (121.0, 0.9598133333333333)
  (122.0, 0.9638733333333334)
  (123.0, 0.9675866666666667)
  (124.0, 0.971)
  (125.0, 0.97376)
  (126.0, 0.9768)
  (127.0, 0.9796266666666666)
  (128.0, 0.9819333333333333)
  (129.0, 0.9838133333333333)
  (130.0, 0.9852466666666667)
  (131.0, 0.9864066666666667)
  (132.0, 0.9878266666666666)
  (133.0, 0.98894)
  (134.0, 0.9900133333333333)
  (135.0, 0.9910466666666666)
  (136.0, 0.9920866666666667)
  (137.0, 0.9928533333333334)
  (138.0, 0.99354)
  (139.0, 0.9944266666666667)
  (140.0, 0.9949333333333333)
  (141.0, 0.9956)
  (142.0, 0.9961666666666666)
  (143.0, 0.9967066666666666)
  (144.0, 0.99726)
  (145.0, 0.9979666666666667)
  (146.0, 0.9984533333333333)
  (147.0, 0.99884)
  (148.0, 0.99936)
  (149.0, 0.9998733333333333)
  (150.0, 1.0)
};

\addplot+[
  pastelBlue, mark = {none}
] coordinates {
  (0.0, 0.0)
  (1.0, 0.0)
  (2.0, 0.0)
  (3.0, 0.0)
  (4.0, 0.00013333333333333334)
  (5.0, 0.0003066666666666667)
  (6.0, 0.0004133333333333333)
  (7.0, 0.0006666666666666666)
  (8.0, 0.0009133333333333334)
  (9.0, 0.0011533333333333333)
  (10.0, 0.00134)
  (11.0, 0.00162)
  (12.0, 0.00176)
  (13.0, 0.00188)
  (14.0, 0.002193333333333333)
  (15.0, 0.0026266666666666665)
  (16.0, 0.0031533333333333335)
  (17.0, 0.0032866666666666665)
  (18.0, 0.0035466666666666667)
  (19.0, 0.003833333333333333)
  (20.0, 0.00422)
  (21.0, 0.00498)
  (22.0, 0.005346666666666666)
  (23.0, 0.005686666666666666)
  (24.0, 0.0064466666666666665)
  (25.0, 0.007173333333333334)
  (26.0, 0.007633333333333333)
  (27.0, 0.008186666666666667)
  (28.0, 0.0087)
  (29.0, 0.009606666666666666)
  (30.0, 0.010426666666666667)
  (31.0, 0.011646666666666666)
  (32.0, 0.013233333333333333)
  (33.0, 0.014153333333333334)
  (34.0, 0.014946666666666667)
  (35.0, 0.016486666666666667)
  (36.0, 0.01782)
  (37.0, 0.018793333333333332)
  (38.0, 0.020286666666666668)
  (39.0, 0.021673333333333333)
  (40.0, 0.022813333333333335)
  (41.0, 0.024486666666666667)
  (42.0, 0.02608)
  (43.0, 0.02794666666666667)
  (44.0, 0.029513333333333332)
  (45.0, 0.031033333333333333)
  (46.0, 0.03324666666666667)
  (47.0, 0.03544666666666667)
  (48.0, 0.03798666666666667)
  (49.0, 0.03997333333333333)
  (50.0, 0.04189333333333333)
  (51.0, 0.04391333333333333)
  (52.0, 0.04601333333333334)
  (53.0, 0.04828)
  (54.0, 0.05114)
  (55.0, 0.05383333333333333)
  (56.0, 0.0571)
  (57.0, 0.05952)
  (58.0, 0.06260666666666667)
  (59.0, 0.066)
  (60.0, 0.06922)
  (61.0, 0.07243333333333334)
  (62.0, 0.07600666666666667)
  (63.0, 0.07942666666666667)
  (64.0, 0.08346666666666666)
  (65.0, 0.08774)
  (66.0, 0.09152666666666667)
  (67.0, 0.09561333333333333)
  (68.0, 0.10025333333333333)
  (69.0, 0.10526)
  (70.0, 0.1101)
  (71.0, 0.11429333333333333)
  (72.0, 0.11896)
  (73.0, 0.12346)
  (74.0, 0.12903333333333333)
  (75.0, 0.13509333333333334)
  (76.0, 0.14248)
  (77.0, 0.14932666666666666)
  (78.0, 0.156)
  (79.0, 0.16479333333333332)
  (80.0, 0.19237333333333334)
  (81.0, 0.2524133333333333)
  (82.0, 0.3030333333333333)
  (83.0, 0.33924)
  (84.0, 0.37371333333333334)
  (85.0, 0.4059333333333333)
  (86.0, 0.4325466666666667)
  (87.0, 0.45776)
  (88.0, 0.48151333333333335)
  (89.0, 0.5047066666666666)
  (90.0, 0.5288133333333334)
  (91.0, 0.55146)
  (92.0, 0.5723466666666667)
  (93.0, 0.5951066666666667)
  (94.0, 0.6202)
  (95.0, 0.63884)
  (96.0, 0.6596666666666666)
  (97.0, 0.6806733333333334)
  (98.0, 0.7038066666666667)
  (99.0, 0.72332)
  (100.0, 0.7421266666666667)
  (101.0, 0.76322)
  (102.0, 0.7836866666666666)
  (103.0, 0.8007733333333333)
  (104.0, 0.8192666666666667)
  (105.0, 0.8378866666666667)
  (106.0, 0.8546266666666666)
  (107.0, 0.8725)
  (108.0, 0.8879133333333333)
  (109.0, 0.9032333333333333)
  (110.0, 0.9185466666666666)
  (111.0, 0.9329266666666667)
  (112.0, 0.942)
  (113.0, 0.9493266666666667)
  (114.0, 0.9553666666666667)
  (115.0, 0.9603333333333334)
  (116.0, 0.9649666666666666)
  (117.0, 0.9688066666666667)
  (118.0, 0.9723533333333333)
  (119.0, 0.9755466666666667)
  (120.0, 0.9781266666666667)
  (121.0, 0.98102)
  (122.0, 0.9831066666666667)
  (123.0, 0.9850333333333333)
  (124.0, 0.9869266666666666)
  (125.0, 0.9886466666666667)
  (126.0, 0.9902333333333333)
  (127.0, 0.9916533333333334)
  (128.0, 0.9927)
  (129.0, 0.9936866666666667)
  (130.0, 0.9943733333333333)
  (131.0, 0.99502)
  (132.0, 0.9956933333333333)
  (133.0, 0.9961733333333334)
  (134.0, 0.9966066666666666)
  (135.0, 0.9970266666666666)
  (136.0, 0.9973733333333333)
  (137.0, 0.99768)
  (138.0, 0.9979333333333333)
  (139.0, 0.9982733333333333)
  (140.0, 0.9985933333333333)
  (141.0, 0.9986533333333333)
  (142.0, 0.9987533333333334)
  (143.0, 0.99898)
  (144.0, 0.9990666666666667)
  (145.0, 0.9993066666666667)
  (146.0, 0.9994666666666666)
  (147.0, 0.9995533333333333)
  (148.0, 0.9997133333333333)
  (149.0, 0.99986)
  (150.0, 1.0)
};

\addplot+[
  pastelRed, mark = {none}
] coordinates {
  (0.0, 0.0)
  (1.0, 0.0)
  (2.0, 0.0)
  (3.0, 6.666666666666667e-6)
  (4.0, 6.666666666666667e-5)
  (5.0, 0.00016)
  (6.0, 0.0002)
  (7.0, 0.0003266666666666667)
  (8.0, 0.0003933333333333333)
  (9.0, 0.0005266666666666667)
  (10.0, 0.0006466666666666667)
  (11.0, 0.0007866666666666666)
  (12.0, 0.00084)
  (13.0, 0.0009266666666666667)
  (14.0, 0.0009866666666666667)
  (15.0, 0.00124)
  (16.0, 0.0014933333333333333)
  (17.0, 0.0016933333333333334)
  (18.0, 0.00186)
  (19.0, 0.0019666666666666665)
  (20.0, 0.0022)
  (21.0, 0.0025666666666666667)
  (22.0, 0.0029333333333333334)
  (23.0, 0.003193333333333333)
  (24.0, 0.0035533333333333333)
  (25.0, 0.004046666666666666)
  (26.0, 0.004213333333333334)
  (27.0, 0.00444)
  (28.0, 0.004866666666666667)
  (29.0, 0.005513333333333334)
  (30.0, 0.006146666666666667)
  (31.0, 0.006546666666666667)
  (32.0, 0.00742)
  (33.0, 0.008233333333333334)
  (34.0, 0.009133333333333334)
  (35.0, 0.010053333333333333)
  (36.0, 0.010706666666666666)
  (37.0, 0.011446666666666667)
  (38.0, 0.0125)
  (39.0, 0.013473333333333334)
  (40.0, 0.014246666666666666)
  (41.0, 0.015193333333333333)
  (42.0, 0.016406666666666667)
  (43.0, 0.017946666666666666)
  (44.0, 0.019153333333333335)
  (45.0, 0.020253333333333335)
  (46.0, 0.021786666666666666)
  (47.0, 0.022893333333333335)
  (48.0, 0.024186666666666665)
  (49.0, 0.025426666666666667)
  (50.0, 0.0267)
  (51.0, 0.027973333333333333)
  (52.0, 0.02954)
  (53.0, 0.03148666666666667)
  (54.0, 0.033613333333333335)
  (55.0, 0.036066666666666664)
  (56.0, 0.038533333333333336)
  (57.0, 0.04032)
  (58.0, 0.04208666666666667)
  (59.0, 0.044526666666666666)
  (60.0, 0.04680666666666667)
  (61.0, 0.04886)
  (62.0, 0.05123333333333333)
  (63.0, 0.053906666666666665)
  (64.0, 0.056306666666666665)
  (65.0, 0.05982)
  (66.0, 0.06277333333333333)
  (67.0, 0.06546666666666667)
  (68.0, 0.06854)
  (69.0, 0.07243333333333334)
  (70.0, 0.07626)
  (71.0, 0.07972666666666667)
  (72.0, 0.08443333333333333)
  (73.0, 0.08872)
  (74.0, 0.0933)
  (75.0, 0.09795333333333334)
  (76.0, 0.10327333333333333)
  (77.0, 0.10860666666666667)
  (78.0, 0.11414)
  (79.0, 0.1213)
  (80.0, 0.14922)
  (81.0, 0.21113333333333334)
  (82.0, 0.26192666666666664)
  (83.0, 0.29929333333333336)
  (84.0, 0.33326)
  (85.0, 0.36702666666666667)
  (86.0, 0.3944533333333333)
  (87.0, 0.42042666666666667)
  (88.0, 0.4454266666666667)
  (89.0, 0.4693866666666667)
  (90.0, 0.49450666666666665)
  (91.0, 0.51746)
  (92.0, 0.5392933333333333)
  (93.0, 0.5621066666666666)
  (94.0, 0.5877733333333334)
  (95.0, 0.60688)
  (96.0, 0.6277133333333333)
  (97.0, 0.6497066666666667)
  (98.0, 0.6740333333333334)
  (99.0, 0.6946866666666667)
  (100.0, 0.7145933333333333)
  (101.0, 0.7373)
  (102.0, 0.7583066666666667)
  (103.0, 0.7762933333333333)
  (104.0, 0.79616)
  (105.0, 0.8160733333333333)
  (106.0, 0.8330733333333333)
  (107.0, 0.8513)
  (108.0, 0.8666733333333333)
  (109.0, 0.88246)
  (110.0, 0.8983933333333334)
  (111.0, 0.91314)
  (112.0, 0.9227466666666667)
  (113.0, 0.93056)
  (114.0, 0.9371333333333334)
  (115.0, 0.9425933333333333)
  (116.0, 0.9478066666666667)
  (117.0, 0.9521866666666666)
  (118.0, 0.9563533333333334)
  (119.0, 0.96024)
  (120.0, 0.9638933333333334)
  (121.0, 0.9674533333333334)
  (122.0, 0.9708733333333334)
  (123.0, 0.97384)
  (124.0, 0.9765333333333334)
  (125.0, 0.9787333333333333)
  (126.0, 0.98146)
  (127.0, 0.9837333333333333)
  (128.0, 0.9855666666666667)
  (129.0, 0.9872533333333333)
  (130.0, 0.9888666666666667)
  (131.0, 0.99004)
  (132.0, 0.9911533333333333)
  (133.0, 0.9921)
  (134.0, 0.9929466666666666)
  (135.0, 0.9938133333333333)
  (136.0, 0.9944466666666667)
  (137.0, 0.9951866666666667)
  (138.0, 0.9957066666666666)
  (139.0, 0.99646)
  (140.0, 0.9970666666666667)
  (141.0, 0.99746)
  (142.0, 0.9977066666666666)
  (143.0, 0.99818)
  (144.0, 0.9984933333333333)
  (145.0, 0.9990133333333333)
  (146.0, 0.9993266666666667)
  (147.0, 0.9994066666666667)
  (148.0, 0.99966)
  (149.0, 0.99986)
  (150.0, 1.0)
};

\addplot+[
  pastelPurple, mark = {none}
] coordinates {
  (0.0, 0.0)
  (1.0, 0.0)
  (2.0, 0.0)
  (3.0, 6.0e-5)
  (4.0, 0.00020666666666666666)
  (5.0, 0.0005133333333333333)
  (6.0, 0.00068)
  (7.0, 0.0010533333333333334)
  (8.0, 0.0012933333333333334)
  (9.0, 0.0015266666666666666)
  (10.0, 0.0017466666666666668)
  (11.0, 0.00196)
  (12.0, 0.002106666666666667)
  (13.0, 0.0022533333333333333)
  (14.0, 0.0024933333333333335)
  (15.0, 0.0031266666666666665)
  (16.0, 0.0036666666666666666)
  (17.0, 0.003986666666666667)
  (18.0, 0.00418)
  (19.0, 0.00436)
  (20.0, 0.00492)
  (21.0, 0.005706666666666666)
  (22.0, 0.00622)
  (23.0, 0.006653333333333333)
  (24.0, 0.00732)
  (25.0, 0.00816)
  (26.0, 0.008453333333333334)
  (27.0, 0.0089)
  (28.0, 0.009766666666666667)
  (29.0, 0.01054)
  (30.0, 0.011766666666666667)
  (31.0, 0.012686666666666667)
  (32.0, 0.014173333333333333)
  (33.0, 0.015486666666666666)
  (34.0, 0.016686666666666666)
  (35.0, 0.01796)
  (36.0, 0.019413333333333335)
  (37.0, 0.02066)
  (38.0, 0.022353333333333333)
  (39.0, 0.023893333333333332)
  (40.0, 0.025413333333333333)
  (41.0, 0.02702)
  (42.0, 0.028833333333333332)
  (43.0, 0.030993333333333335)
  (44.0, 0.0329)
  (45.0, 0.03472)
  (46.0, 0.03716)
  (47.0, 0.038906666666666666)
  (48.0, 0.040793333333333334)
  (49.0, 0.042493333333333334)
  (50.0, 0.044353333333333335)
  (51.0, 0.046106666666666664)
  (52.0, 0.04831333333333333)
  (53.0, 0.05125333333333333)
  (54.0, 0.0547)
  (55.0, 0.058)
  (56.0, 0.06148)
  (57.0, 0.06392666666666667)
  (58.0, 0.06651333333333333)
  (59.0, 0.07005333333333333)
  (60.0, 0.07349333333333333)
  (61.0, 0.07656)
  (62.0, 0.08044)
  (63.0, 0.08422666666666667)
  (64.0, 0.08804)
  (65.0, 0.0919)
  (66.0, 0.09621333333333333)
  (67.0, 0.10088666666666667)
  (68.0, 0.10556)
  (69.0, 0.11062666666666666)
  (70.0, 0.116)
  (71.0, 0.12101333333333333)
  (72.0, 0.12616)
  (73.0, 0.13196)
  (74.0, 0.13815333333333332)
  (75.0, 0.14412)
  (76.0, 0.15158)
  (77.0, 0.15846666666666667)
  (78.0, 0.16569333333333333)
  (79.0, 0.17457333333333333)
  (80.0, 0.20053333333333334)
  (81.0, 0.25992666666666664)
  (82.0, 0.30986)
  (83.0, 0.3459466666666667)
  (84.0, 0.38083333333333336)
  (85.0, 0.4137266666666667)
  (86.0, 0.4406)
  (87.0, 0.46588666666666667)
  (88.0, 0.48940666666666666)
  (89.0, 0.5132933333333334)
  (90.0, 0.5377666666666666)
  (91.0, 0.56082)
  (92.0, 0.5820533333333333)
  (93.0, 0.6047666666666667)
  (94.0, 0.6301333333333333)
  (95.0, 0.6486)
  (96.0, 0.6686133333333333)
  (97.0, 0.6893)
  (98.0, 0.7115866666666667)
  (99.0, 0.73104)
  (100.0, 0.7496733333333333)
  (101.0, 0.7703866666666667)
  (102.0, 0.7901933333333333)
  (103.0, 0.80684)
  (104.0, 0.8245)
  (105.0, 0.8429533333333333)
  (106.0, 0.8596666666666667)
  (107.0, 0.87742)
  (108.0, 0.8926533333333333)
  (109.0, 0.9074666666666666)
  (110.0, 0.9221333333333334)
  (111.0, 0.9359933333333333)
  (112.0, 0.9450666666666667)
  (113.0, 0.95208)
  (114.0, 0.9577)
  (115.0, 0.9625333333333334)
  (116.0, 0.9668666666666667)
  (117.0, 0.9704066666666666)
  (118.0, 0.9737933333333333)
  (119.0, 0.9769266666666667)
  (120.0, 0.9795266666666667)
  (121.0, 0.9822)
  (122.0, 0.9841533333333333)
  (123.0, 0.98602)
  (124.0, 0.9876133333333333)
  (125.0, 0.9892133333333333)
  (126.0, 0.9907466666666667)
  (127.0, 0.9919533333333334)
  (128.0, 0.993)
  (129.0, 0.9942133333333333)
  (130.0, 0.9948866666666667)
  (131.0, 0.9953733333333333)
  (132.0, 0.9959333333333333)
  (133.0, 0.99642)
  (134.0, 0.99688)
  (135.0, 0.9973733333333333)
  (136.0, 0.9976466666666667)
  (137.0, 0.9978333333333333)
  (138.0, 0.9981066666666667)
  (139.0, 0.9982933333333334)
  (140.0, 0.9985533333333333)
  (141.0, 0.9985866666666666)
  (142.0, 0.9987133333333333)
  (143.0, 0.99904)
  (144.0, 0.99918)
  (145.0, 0.9994933333333333)
  (146.0, 0.9996133333333334)
  (147.0, 0.99974)
  (148.0, 0.9998866666666667)
  (149.0, 0.99992)
  (150.0, 1.0)
};
\legend{Baseline, Risk Loss, Risk Data, Both};

\end{groupplot}

\end{tikzpicture}

%% file: daa_example.tex
\begin{tikzpicture}[]

\begin{axis}[
  height = {9.8cm},
  ylabel = {North (m)},
  axis equal image = {true},
  xlabel = {East (m)},
  at = {(-18,-75)},
  xtick = {500,1000},
  ymin = {0},
]

\input{ground_track}

\end{axis}

\begin{groupplot}[group style={horizontal sep = 0.5cm, vertical sep = 0.1cm, group size=1 by 2}]

\input{altitude_risk_profile}

\end{groupplot}

\node at (4.2, 4.5) {\includegraphics[width=9.1cm]{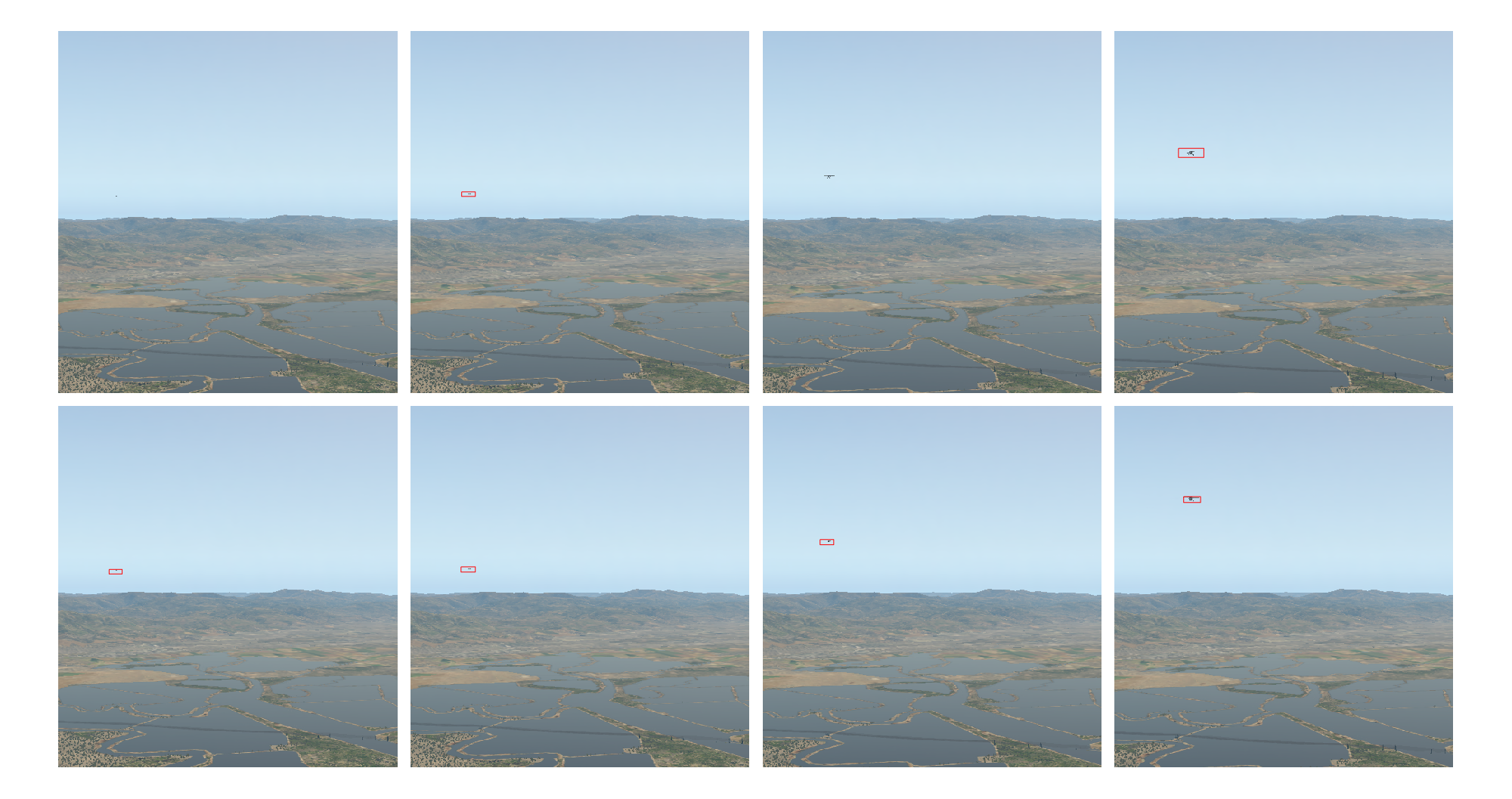}};

\node[rotate=90] at (-0.4, 5.6) {Baseline};
\node[rotate=90] at (-0.4, 3.35) {Risk Sensitive};

\draw[->, opacity=0.75] (1, 2.3) -- (4.8, 1.45);
\draw[->, opacity=0.75] (3.1, 2.3) -- (5.3, 1.45);
\draw[->, opacity=0.75] (4.2, 2.3) -- (6.5, 1.45);
\draw[->, opacity=0.75] (7.4, 2.3) -- (7.37, 1.45);

\node[red] at (-3, 6) {\scriptsize intruder};
\node[gray] at (-2.5, -1) {\scriptsize ownship};

\end{tikzpicture}

%% file: daa_pol.tex
\begin{tikzpicture}[]
\begin{axis}[
  ylabel = {$h$ (m)},
  xlabel = {$\tau$ (s)},
  enlargelimits = false,
  axis on top,
  height={6cm}
]

\addplot[
  point meta min = 1,
  point meta max = 3
] graphics[
  xmin = 0,
  xmax = 40,
  ymin = -300,
  ymax = 300
] {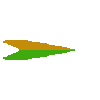};

\end{axis}
\end{tikzpicture}

%% file: daa_em.tex
\begin{tikzpicture}[]
\begin{axis}[
  ylabel = {$h$ (m)},
  title = {Probability of Detection},
  xlabel = {$\tau$ (s)},
  enlargelimits = false,
  height={6cm},
  axis on top,
  colormap={mycolormap}{ rgb(0cm)=(0.267004,0.004874,0.329415) rgb(1cm)=(0.26851,0.009605,0.335427) rgb(2cm)=(0.269944,0.014625,0.341379) rgb(3cm)=(0.271305,0.019942,0.347269) rgb(4cm)=(0.272594,0.025563,0.353093) rgb(5cm)=(0.273809,0.031497,0.358853) rgb(6cm)=(0.274952,0.037752,0.364543) rgb(7cm)=(0.276022,0.044167,0.370164) rgb(8cm)=(0.277018,0.050344,0.375715) rgb(9cm)=(0.277941,0.056324,0.381191) rgb(10cm)=(0.278791,0.062145,0.386592) rgb(11cm)=(0.279566,0.067836,0.391917) rgb(12cm)=(0.280267,0.073417,0.397163) rgb(13cm)=(0.280894,0.078907,0.402329) rgb(14cm)=(0.281446,0.08432,0.407414) rgb(15cm)=(0.281924,0.089666,0.412415) rgb(16cm)=(0.282327,0.094955,0.417331) rgb(17cm)=(0.282656,0.100196,0.42216) rgb(18cm)=(0.28291,0.105393,0.426902) rgb(19cm)=(0.283091,0.110553,0.431554) rgb(20cm)=(0.283197,0.11568,0.436115) rgb(21cm)=(0.283229,0.120777,0.440584) rgb(22cm)=(0.283187,0.125848,0.44496) rgb(23cm)=(0.283072,0.130895,0.449241) rgb(24cm)=(0.282884,0.13592,0.453427) rgb(25cm)=(0.282623,0.140926,0.457517) rgb(26cm)=(0.28229,0.145912,0.46151) rgb(27cm)=(0.281887,0.150881,0.465405) rgb(28cm)=(0.281412,0.155834,0.469201) rgb(29cm)=(0.280868,0.160771,0.472899) rgb(30cm)=(0.280255,0.165693,0.476498) rgb(31cm)=(0.279574,0.170599,0.479997) rgb(32cm)=(0.278826,0.17549,0.483397) rgb(33cm)=(0.278012,0.180367,0.486697) rgb(34cm)=(0.277134,0.185228,0.489898) rgb(35cm)=(0.276194,0.190074,0.493001) rgb(36cm)=(0.275191,0.194905,0.496005) rgb(37cm)=(0.274128,0.199721,0.498911) rgb(38cm)=(0.273006,0.20452,0.501721) rgb(39cm)=(0.271828,0.209303,0.504434) rgb(40cm)=(0.270595,0.214069,0.507052) rgb(41cm)=(0.269308,0.218818,0.509577) rgb(42cm)=(0.267968,0.223549,0.512008) rgb(43cm)=(0.26658,0.228262,0.514349) rgb(44cm)=(0.265145,0.232956,0.516599) rgb(45cm)=(0.263663,0.237631,0.518762) rgb(46cm)=(0.262138,0.242286,0.520837) rgb(47cm)=(0.260571,0.246922,0.522828) rgb(48cm)=(0.258965,0.251537,0.524736) rgb(49cm)=(0.257322,0.25613,0.526563) rgb(50cm)=(0.255645,0.260703,0.528312) rgb(51cm)=(0.253935,0.265254,0.529983) rgb(52cm)=(0.252194,0.269783,0.531579) rgb(53cm)=(0.250425,0.27429,0.533103) rgb(54cm)=(0.248629,0.278775,0.534556) rgb(55cm)=(0.246811,0.283237,0.535941) rgb(56cm)=(0.244972,0.287675,0.53726) rgb(57cm)=(0.243113,0.292092,0.538516) rgb(58cm)=(0.241237,0.296485,0.539709) rgb(59cm)=(0.239346,0.300855,0.540844) rgb(60cm)=(0.237441,0.305202,0.541921) rgb(61cm)=(0.235526,0.309527,0.542944) rgb(62cm)=(0.233603,0.313828,0.543914) rgb(63cm)=(0.231674,0.318106,0.544834) rgb(64cm)=(0.229739,0.322361,0.545706) rgb(65cm)=(0.227802,0.326594,0.546532) rgb(66cm)=(0.225863,0.330805,0.547314) rgb(67cm)=(0.223925,0.334994,0.548053) rgb(68cm)=(0.221989,0.339161,0.548752) rgb(69cm)=(0.220057,0.343307,0.549413) rgb(70cm)=(0.21813,0.347432,0.550038) rgb(71cm)=(0.21621,0.351535,0.550627) rgb(72cm)=(0.214298,0.355619,0.551184) rgb(73cm)=(0.212395,0.359683,0.55171) rgb(74cm)=(0.210503,0.363727,0.552206) rgb(75cm)=(0.208623,0.367752,0.552675) rgb(76cm)=(0.206756,0.371758,0.553117) rgb(77cm)=(0.204903,0.375746,0.553533) rgb(78cm)=(0.203063,0.379716,0.553925) rgb(79cm)=(0.201239,0.38367,0.554294) rgb(80cm)=(0.19943,0.387607,0.554642) rgb(81cm)=(0.197636,0.391528,0.554969) rgb(82cm)=(0.19586,0.395433,0.555276) rgb(83cm)=(0.1941,0.399323,0.555565) rgb(84cm)=(0.192357,0.403199,0.555836) rgb(85cm)=(0.190631,0.407061,0.556089) rgb(86cm)=(0.188923,0.41091,0.556326) rgb(87cm)=(0.187231,0.414746,0.556547) rgb(88cm)=(0.185556,0.41857,0.556753) rgb(89cm)=(0.183898,0.422383,0.556944) rgb(90cm)=(0.182256,0.426184,0.55712) rgb(91cm)=(0.180629,0.429975,0.557282) rgb(92cm)=(0.179019,0.433756,0.55743) rgb(93cm)=(0.177423,0.437527,0.557565) rgb(94cm)=(0.175841,0.44129,0.557685) rgb(95cm)=(0.174274,0.445044,0.557792) rgb(96cm)=(0.172719,0.448791,0.557885) rgb(97cm)=(0.171176,0.45253,0.557965) rgb(98cm)=(0.169646,0.456262,0.55803) rgb(99cm)=(0.168126,0.459988,0.558082) rgb(100cm)=(0.166617,0.463708,0.558119) rgb(101cm)=(0.165117,0.467423,0.558141) rgb(102cm)=(0.163625,0.471133,0.558148) rgb(103cm)=(0.162142,0.474838,0.55814) rgb(104cm)=(0.160665,0.47854,0.558115) rgb(105cm)=(0.159194,0.482237,0.558073) rgb(106cm)=(0.157729,0.485932,0.558013) rgb(107cm)=(0.15627,0.489624,0.557936) rgb(108cm)=(0.154815,0.493313,0.55784) rgb(109cm)=(0.153364,0.497,0.557724) rgb(110cm)=(0.151918,0.500685,0.557587) rgb(111cm)=(0.150476,0.504369,0.55743) rgb(112cm)=(0.149039,0.508051,0.55725) rgb(113cm)=(0.147607,0.511733,0.557049) rgb(114cm)=(0.14618,0.515413,0.556823) rgb(115cm)=(0.144759,0.519093,0.556572) rgb(116cm)=(0.143343,0.522773,0.556295) rgb(117cm)=(0.141935,0.526453,0.555991) rgb(118cm)=(0.140536,0.530132,0.555659) rgb(119cm)=(0.139147,0.533812,0.555298) rgb(120cm)=(0.13777,0.537492,0.554906) rgb(121cm)=(0.136408,0.541173,0.554483) rgb(122cm)=(0.135066,0.544853,0.554029) rgb(123cm)=(0.133743,0.548535,0.553541) rgb(124cm)=(0.132444,0.552216,0.553018) rgb(125cm)=(0.131172,0.555899,0.552459) rgb(126cm)=(0.129933,0.559582,0.551864) rgb(127cm)=(0.128729,0.563265,0.551229) rgb(128cm)=(0.127568,0.566949,0.550556) rgb(129cm)=(0.126453,0.570633,0.549841) rgb(130cm)=(0.125394,0.574318,0.549086) rgb(131cm)=(0.124395,0.578002,0.548287) rgb(132cm)=(0.123463,0.581687,0.547445) rgb(133cm)=(0.122606,0.585371,0.546557) rgb(134cm)=(0.121831,0.589055,0.545623) rgb(135cm)=(0.121148,0.592739,0.544641) rgb(136cm)=(0.120565,0.596422,0.543611) rgb(137cm)=(0.120092,0.600104,0.54253) rgb(138cm)=(0.119738,0.603785,0.5414) rgb(139cm)=(0.119512,0.607464,0.540218) rgb(140cm)=(0.119423,0.611141,0.538982) rgb(141cm)=(0.119483,0.614817,0.537692) rgb(142cm)=(0.119699,0.61849,0.536347) rgb(143cm)=(0.120081,0.622161,0.534946) rgb(144cm)=(0.120638,0.625828,0.533488) rgb(145cm)=(0.12138,0.629492,0.531973) rgb(146cm)=(0.122312,0.633153,0.530398) rgb(147cm)=(0.123444,0.636809,0.528763) rgb(148cm)=(0.12478,0.640461,0.527068) rgb(149cm)=(0.126326,0.644107,0.525311) rgb(150cm)=(0.128087,0.647749,0.523491) rgb(151cm)=(0.130067,0.651384,0.521608) rgb(152cm)=(0.132268,0.655014,0.519661) rgb(153cm)=(0.134692,0.658636,0.517649) rgb(154cm)=(0.137339,0.662252,0.515571) rgb(155cm)=(0.14021,0.665859,0.513427) rgb(156cm)=(0.143303,0.669459,0.511215) rgb(157cm)=(0.146616,0.67305,0.508936) rgb(158cm)=(0.150148,0.676631,0.506589) rgb(159cm)=(0.153894,0.680203,0.504172) rgb(160cm)=(0.157851,0.683765,0.501686) rgb(161cm)=(0.162016,0.687316,0.499129) rgb(162cm)=(0.166383,0.690856,0.496502) rgb(163cm)=(0.170948,0.694384,0.493803) rgb(164cm)=(0.175707,0.6979,0.491033) rgb(165cm)=(0.180653,0.701402,0.488189) rgb(166cm)=(0.185783,0.704891,0.485273) rgb(167cm)=(0.19109,0.708366,0.482284) rgb(168cm)=(0.196571,0.711827,0.479221) rgb(169cm)=(0.202219,0.715272,0.476084) rgb(170cm)=(0.20803,0.718701,0.472873) rgb(171cm)=(0.214,0.722114,0.469588) rgb(172cm)=(0.220124,0.725509,0.466226) rgb(173cm)=(0.226397,0.728888,0.462789) rgb(174cm)=(0.232815,0.732247,0.459277) rgb(175cm)=(0.239374,0.735588,0.455688) rgb(176cm)=(0.24607,0.73891,0.452024) rgb(177cm)=(0.252899,0.742211,0.448284) rgb(178cm)=(0.259857,0.745492,0.444467) rgb(179cm)=(0.266941,0.748751,0.440573) rgb(180cm)=(0.274149,0.751988,0.436601) rgb(181cm)=(0.281477,0.755203,0.432552) rgb(182cm)=(0.288921,0.758394,0.428426) rgb(183cm)=(0.296479,0.761561,0.424223) rgb(184cm)=(0.304148,0.764704,0.419943) rgb(185cm)=(0.311925,0.767822,0.415586) rgb(186cm)=(0.319809,0.770914,0.411152) rgb(187cm)=(0.327796,0.77398,0.40664) rgb(188cm)=(0.335885,0.777018,0.402049) rgb(189cm)=(0.344074,0.780029,0.397381) rgb(190cm)=(0.35236,0.783011,0.392636) rgb(191cm)=(0.360741,0.785964,0.387814) rgb(192cm)=(0.369214,0.788888,0.382914) rgb(193cm)=(0.377779,0.791781,0.377939) rgb(194cm)=(0.386433,0.794644,0.372886) rgb(195cm)=(0.395174,0.797475,0.367757) rgb(196cm)=(0.404001,0.800275,0.362552) rgb(197cm)=(0.412913,0.803041,0.357269) rgb(198cm)=(0.421908,0.805774,0.35191) rgb(199cm)=(0.430983,0.808473,0.346476) rgb(200cm)=(0.440137,0.811138,0.340967) rgb(201cm)=(0.449368,0.813768,0.335384) rgb(202cm)=(0.458674,0.816363,0.329727) rgb(203cm)=(0.468053,0.818921,0.323998) rgb(204cm)=(0.477504,0.821444,0.318195) rgb(205cm)=(0.487026,0.823929,0.312321) rgb(206cm)=(0.496615,0.826376,0.306377) rgb(207cm)=(0.506271,0.828786,0.300362) rgb(208cm)=(0.515992,0.831158,0.294279) rgb(209cm)=(0.525776,0.833491,0.288127) rgb(210cm)=(0.535621,0.835785,0.281908) rgb(211cm)=(0.545524,0.838039,0.275626) rgb(212cm)=(0.555484,0.840254,0.269281) rgb(213cm)=(0.565498,0.84243,0.262877) rgb(214cm)=(0.575563,0.844566,0.256415) rgb(215cm)=(0.585678,0.846661,0.249897) rgb(216cm)=(0.595839,0.848717,0.243329) rgb(217cm)=(0.606045,0.850733,0.236712) rgb(218cm)=(0.616293,0.852709,0.230052) rgb(219cm)=(0.626579,0.854645,0.223353) rgb(220cm)=(0.636902,0.856542,0.21662) rgb(221cm)=(0.647257,0.8584,0.209861) rgb(222cm)=(0.657642,0.860219,0.203082) rgb(223cm)=(0.668054,0.861999,0.196293) rgb(224cm)=(0.678489,0.863742,0.189503) rgb(225cm)=(0.688944,0.865448,0.182725) rgb(226cm)=(0.699415,0.867117,0.175971) rgb(227cm)=(0.709898,0.868751,0.169257) rgb(228cm)=(0.720391,0.87035,0.162603) rgb(229cm)=(0.730889,0.871916,0.156029) rgb(230cm)=(0.741388,0.873449,0.149561) rgb(231cm)=(0.751884,0.874951,0.143228) rgb(232cm)=(0.762373,0.876424,0.137064) rgb(233cm)=(0.772852,0.877868,0.131109) rgb(234cm)=(0.783315,0.879285,0.125405) rgb(235cm)=(0.79376,0.880678,0.120005) rgb(236cm)=(0.804182,0.882046,0.114965) rgb(237cm)=(0.814576,0.883393,0.110347) rgb(238cm)=(0.82494,0.88472,0.106217) rgb(239cm)=(0.83527,0.886029,0.102646) rgb(240cm)=(0.845561,0.887322,0.099702) rgb(241cm)=(0.85581,0.888601,0.097452) rgb(242cm)=(0.866013,0.889868,0.095953) rgb(243cm)=(0.876168,0.891125,0.09525) rgb(244cm)=(0.886271,0.892374,0.095374) rgb(245cm)=(0.89632,0.893616,0.096335) rgb(246cm)=(0.906311,0.894855,0.098125) rgb(247cm)=(0.916242,0.896091,0.100717) rgb(248cm)=(0.926106,0.89733,0.104071) rgb(249cm)=(0.935904,0.89857,0.108131) rgb(250cm)=(0.945636,0.899815,0.112838) rgb(251cm)=(0.9553,0.901065,0.118128) rgb(252cm)=(0.964894,0.902323,0.123941) rgb(253cm)=(0.974417,0.90359,0.130215) rgb(254cm)=(0.983868,0.904867,0.136897) rgb(255cm)=(0.993248,0.906157,0.143936) },
  colorbar
]

\addplot[
  point meta min = 0,
  point meta max = 1
] graphics[
  xmin = 0,
  xmax = 40,
  ymin = -300,
  ymax = 300
] {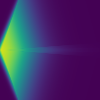};

\end{axis}
\end{tikzpicture}